\documentclass{article}

\usepackage{amsmath,amsthm,amssymb,amsfonts}
\usepackage{graphicx}
\usepackage{arxiv}

\usepackage[utf8]{inputenc} 
\usepackage[T1]{fontenc}    
\usepackage{hyperref}       
\usepackage{url}            
\usepackage{booktabs}       
\usepackage{amsfonts}       
\usepackage{nicefrac}       
\usepackage{microtype}      
\usepackage{lipsum}

\usepackage[utf8]{inputenc} 
\usepackage[T1]{fontenc}    
\usepackage{hyperref}       
\usepackage{url}            
\usepackage{booktabs}       
\usepackage{amsfonts}       
\usepackage{nicefrac}       
\usepackage{microtype}      
\usepackage{xcolor}         
\usepackage{multirow}
\usepackage{caption}
\usepackage{subcaption}
\usepackage{graphicx}
\usepackage{amsfonts, amsmath, amssymb} 
\usepackage{algorithm}
\usepackage{algorithmic}

\usepackage{wrapfig}
\usepackage{tikz}
\usepackage{array}

\newcommand{\D}{\mathcal{D}}
\newcommand{\R}{\mathbb{R}}
\newcommand{\M}{\mathcal{M}}
\newcommand{\GN}{\mathcal{N}}

\DeclareMathOperator{\Clip}{Clip}
\DeclareMathOperator{\Enc}{Enc}
\DeclareMathOperator{\Dec}{Dec}

\DeclareMathOperator{\Noise}{Noise}
\DeclareMathOperator{\Aggre}{Aggregation}

\theoremstyle{plain}
\newtheorem{definition}{Definition}[section]
\newtheorem{theorem}{Theorem}[section]
\newtheorem{lemma}{Lemma}[section]

\theoremstyle{remark}

\usepackage{newfloat}
\usepackage{listings}

\newcommand{\modelname}{\textbf{DR-Encoder}}

\usepackage{enumitem}

\title{DR-Encoder: Encode Low-rank Gradients with Random Prior for Large Language Models Differentially Privately~\thanks{This work is accepted by AAAI-25 AI Alignment Track.}}

\author{
  Huiwen Wu \\
  Zhejiang Laboratory\\
  Hangzhou, Zhejiang, China \\
  \texttt{whw@zhejianglab.org} \\
   \And
    Deyi Zhang \\
    Zhejiang Laboratory\\
    Hangzhou, Zhejiang, China \\
    \texttt{xiaohan@zhejianglab.org} \\
   \And
    Xiaohan Li \\
    Zhejiang Laboratory\\
    Hangzhou, Zhejiang, China \\
    \texttt{xiaohan@zhejianglab.org} \\
   \And
    Xiaogang Xu~ \thanks{corresponding author} \\
    The Chinese University of Hong Kong \\
    Hong Kong, China \\
    \texttt{xiaogangxu00@gmail.com} \\
   \And
    Jiafei Wu~ \thanks{corresponding author} \\
    Zhejiang Laboratory\\
    Hangzhou, Zhejiang, China \\
    \texttt{wujiafei@zhejianglab.org} \\
   \And
    Zhe Liu \\
    Zhejiang Laboratory\\
    Hangzhou, Zhejiang, China \\
    \texttt{zhe.liu@zhejianglab.org} \\
}

\begin{document}
\maketitle

\begin{abstract}

The emergence of the Large Language Model (LLM) has shown their superiority in a wide range of disciplines, including language understanding and translation, relational logic reasoning, and even partial differential equations solving.
The transformer is the pervasive backbone architecture for the foundation model construction. 
It is vital to research how to adjust the Transformer architecture to achieve an end-to-end privacy guarantee in LLM fine-tuning. 
In this paper, we investigate three potential information leakage during a federated fine-tuning procedure for LLM (FedLLM). 
Based on the potential information leakage, we provide an end-to-end privacy guarantee solution for FedLLM by inserting two-stage randomness. 
The first stage is to train a gradient auto-encoder with a Gaussian random prior based on the statistical information of the gradients generated by local clients. 
The second stage is to fine-tune the overall LLM with a differential privacy guarantee by adopting appropriate Gaussian noises. 
We show the efficiency and accuracy gains of our proposed method with several foundation models and two popular evaluation benchmarks. 
Furthermore, we present a comprehensive privacy analysis with Gaussian Differential Privacy (GDP) and Renyi Differential Privacy (RDP). 
\end{abstract}

\section{Introduction}

Large language models (LLMs) have demonstrated their strong capabilities in real-life applications, such as language understanding~\cite{karanikolas2023large}, mathematical reasoning~\cite{imani2023mathprompter}, and even differential equation solving~\cite{herde2024poseidon}. LLM-based applications provide a variety of convenient tools, including chatbots, virtual assistants, article writing, creative writing, and translation services. However, the extensive use of LLMs in daily work also poses a significant risk of unintentional leakage of personal information.

One typical approach to address privacy concerns in daily LLM usage is to use differential privacy during model training, achieved by adding extra Gaussian noise to the training data or intermediate gradients. However, these methods may lose their effectiveness when dealing with extensive input data and billions of model parameters in LLMs. This paper introduces a practical differential private gradient descent for fine-tuning pre-trained LLMs in a parallel manner with a pre-trained AutoEncoder to encode gradients based on a Gaussian prior. The aim is to achieve differential privacy at a client-level while preserving utility when using LLMs in downstream tasks.

In this paper, we propose a learning-based random prior gradient encoding method for federated LLM, called \modelname.
In our design of the pretraining AutoEncoder with a random prior, intermediate gradients are initially collected using low-rank decomposition (LoRA~\cite{hu2021lora}) during individual training sessions. Afterward, we calculate the mean and variance from the low-rank gradients collected for each layer and training epoch. Subsequently, we create synthetic gradient data derived from the Gaussian distribution with the previously determined mean and standard deviations. This synthetic gradient-shaped data are then employed to train the Auto-Encoder for gradient compression and decompression. 
Next, we illustrate the process when applying the Auto-Encoder to fine-tune a Large Language Model (LLM). The first step involves placing the encoder and decoder on the client side. During federated LLM fine-tuning, original gradients are first perturbed, and then compressed into a subspace representation. This subspace gradient feature is transmitted to the global server, which then decompresses the gradients back into the LoRA shape and aggregates them. After aggregating the local gradients, the server sends back the aggregated gradients, which the client uses for the local gradient descent. 

Compared to current methods that employ learning-based gradient compression~\cite{wu2024cgfedllmcompressgradientsfederated,lin2017deep,li2019end,abrahamyan2021learned}, our proposed method significantly enhances the privacy of the associated contributors. First, the gradient data used for training the AutoEncoder is synthesized from data generated using a Gaussian distribution rather than the real gradients, ensuring a strong privacy guarantee for the system. We rely solely on statistical information of gradients per layer and epoch for pre-training the AutoEncoder. Second, we apply a differential privacy mechanism to compressed gradients to maintain client-level differential privacy in the Federated Learning (FL) training setting. Finally, we offer accurate privacy for the entire federated system by considering randomness in client selection.

\begin{itemize}
\item We propose a novel compression training strategy for FL based on the mean and standard deviation gradients layer-wise. Experiments verifies the utility of the AutoEncoder trained via the synthetic data. 
\item A new differential privacy mechanism is devised to guarantee the privacy of local LoRA gradients. 
\item We adopt the Gaussian Differential Privacy and the Renyi Differential Privacy for a comprehensive privacy analysis both theoretically and numerically. 
\end{itemize}

\section{Related Work}

\subsection{Differential Privacy for LLMs}
EW-Tune~\cite{10031034} presents a differential privacy (DP) framework for fine-tuning LLMs using an edgewise accountant. This approach ensures finite sample privacy guarantees through perturbation applied to the low-rank decomposition of the gradient matrix.
The authors in \cite{shi2022just} introduce a technique named Just Fine-tune Twice (JFT) that aims to achieve selective differential privacy (SDP) in large language models (LLMs) through two protective layers. The first layer, a low contextual detector, secures named entities, proper nouns, pronouns, and sentence components like subjects and objects, while the second layer, a high contextual detector, censors verbs to further enhance privacy.
Whispered Tuning~\cite{singh2024whispered} is a multifaceted approach that integrates redaction of personally identifiable information, differential privacy techniques, and output filtering to improve privacy preservation in LLM.
Split-N-Denoise~\cite{mai2023split} is a system crafted to protect user data privacy during the inference stage of large language models (LLM) by leveraging local differential privacy (LDP). It provides the user with a Transformer-based denoising model pre-trained on the server using public datasets and artificial noise.
In~\cite{charles2024fine}, research study two variants of DP-SGD were investigated in the research study with sampling at the sample level and gradient clipping per sample to achieve differential privacy at the sample level and user-level sampling with clipping of gradient per user to achieve differential privacy at the user level. 

\subsection{Parameter Efficient Fine-tuning (PEFT)}
Mixout approaches~\cite{lee2019mixout} integrate the standard network with the dropout network utilizing a specified probability. 
LoRA methods~\cite{hu2021lora,liu2024dora,dettmers2024qlora} decompose the gradient matrix and reconstruct it by multiplying the low-rank matrices.
Adapter-based methods~\cite{karimi2021compacter,mahabadi2021parameter} introduce an additional adapter layer within the transformer layer, altering the network architecture.
MagPruning methods~\cite{han2015deep,han2015learning,lagunas2021block} follow the principle that large weights are more important. By filtering out small weights in absolute values, it tunes the parameters with large absolute values only. 
DiffPruning~\cite{mallya2018piggyback,guo2021parameter} uses a Bernoulli random variable to represent the mask selection process and learns this variable through reparameterization methods. 
Child-Pruning~\cite{xu2021raise,mostafa2019parameter} trains in the full parameter space and calculates the projected mask to find the child network. 
In~\cite{fu2023effectiveness}, the authors provide a unified sparse fine-tuning model containing random approaches, rule-based approaches, and projection-based approaches. Based on the proposed unified sparse fine-tuned model, it further provides comprehensive theoretical analysis for fine-tuning methods. 
%
%

\section{Methodology}
\subsection{Privacy Goal}
The goal of our methods is to provide an end-to-end privacy guarantee for the gradient compression procedure in federated learning.
We divide the whole gradients compress into two procedures. One is the pre-training of AutoEncoder to acquire the encoder and decoder for gradients compression. The second is federated fine-tuning with clients equipping encoder and server equipping with decoder.
In the aforementioned two stages, there are several chances of information leakage. 
Firstly, when using local training gradients as input for AutoEncoder, the collection and transmission of the local gradients of the clients leak the sensitive information of the clients through the reconstruction attacker from the gradients to the original data~\cite{petrov2024dagerexactgradientinversion} and the membership inference from the gradients~\cite{feng2024exposingprivacygapsmembership,maini2024llmdatasetinferencedid, wei2024provingmembershipllmpretraining}. 
Moreover, as we record the local training gradients as the input for training the AutoEncoder, the local data information is condensed into the models' weights of AutoEncoder. There is a chance to infer the original data information from the trained AutoEncoder, which are called model inversion attacks~\cite{10.1145/2810103.2813677}. 
We took the following steps to achieve the end-to-end privacy guarantee. 
\begin{itemize}
\item Instead of transimission of the exact gradients from client to server, we transmit the stastics of gradients only. And we use the statistics to generate synthetic gradients for AutoEncoder pre-training. 
\item In the fine-tuning stage, instead of transmitting the exact gradients from client to server, we adopt differential privacy on the local gradients.
\item A rigorous analysis of the privacy cost is presented to validate the privacy leakage in the entire federated system. 
\end{itemize}

\subsection{Pre-Training with Random Prior}

In this section, we illustrate the process of training an AutoEncoder to grasp the statistical properties of the training gradients.
Inspired by work on training deep neural networks for gradient compression~\cite{li2019end,wu2024cgfedllmcompressgradientsfederated} and noise reduction~\cite{mai2023split} in conventional ML and LLM, we introduce a novel method to train the AutoEncoder by exclusively sharing the statistical details of the training gradients during the actual training phase.
Thus, protecting the original gradient information during AutoEncoder training is crucial. To achieve this goal, we collect only the statistical information of the gradients, including the mean and standard deviation for each layer and epoch, and send them to the server. The information collected can be stored in the form $[\mathbf{m_{*,i}^t}, \mathbf{s_{*,i}^t}]_i^t,$ where $*$ denotes the low-rank parts of $\mathbf{A}$ or $\mathbf{B}$, $i$ represents the layer index and $t$ indicates the epoch index.
We adopt a dynamical way to compute the mean and standard deviation of local gradients by layer and epoch, which follows steps 1 to 4 on the client side as shown in Algorithm~\ref{alg:aux_params}. 
The hyperparameters $\beta_1, \beta_2, h_1, h_2$ are small scalars.
In our experiments, we use $\beta_1 = 0.99, \beta_2 = 0.9, h_1 = 10^{-5}$ and $h_2 = 10^{-3}$.
When server receiving the collected mean and standard deviation, it generates synthetic gradients with Gaussian distribution. 
The synthetic gradients are in the form 
$
[\mathbf{\hat{G}_i^t}] = [\mathbf{\hat{A}_i^t}, \mathbf{\hat{B}_i^t}], \mathbf{\hat{A}_i^t} \sim \GN(\mathbf{m_{A,i}^t},\mathbf{s_i^t}), \mathbf{B_i^t} \sim \GN(\mathbf{m_{B,i}^t}, \mathbf{s_{B,i}^t}). 
$
The server uses synthetic gradients to train the AutoEncoder with loss of $\ell_2$ reconstruction. 
Once the server completes training the AutoEncoder, it separates the AutoEncoder into an encoder and a decoder, dispatching the encoder to all clients while retaining the decoder exclusively.
We present the training details in Algorithm~\ref{alg:aux_params}.

\begin{algorithm}[t]
\textbf{Input:} 
$[\mathbf{A}_i^t, \mathbf{B}_i^t]_{i \in [M], t \in [N]}$: Gradients tensor of client $i$ at communication step $t$; 
hyper-parameters $\beta_1, \beta_2, h_1, h_2$;
\\
\textbf{Output} 
$\mathbf{Enc}$: the pretrained gradients encoder; 
$\mathbf{Dec}$: the pretrained gradients decoder;   \\
\textbf{Client side:}\\
\textbf{for each client $i \in [M]$}: 
\begin{algorithmic}[1]
{
        		\STATE \textbf{Dynamically estimate mean epoch-wise:} 
		$$
		\mathbf{m^t_{A,i+1}}  = \beta_1 \mathbf{m^t_{A,i}} + \left( 1 - \beta_1 \right) \mathbf{\mathbf{A}_i^t}; 
		$$
		\STATE \textbf{Dynamically estimate variance:}
		$$
		\mathbf{v_{i+1}}  = \min \left( \max \left(  \| \mathbf{A_i^t} - \mathbf{m^t_{A, i+1}} \|^2 , h_1 \right)  h_2 \right);
		$$
		\STATE \textbf{Update estimate standard deviation:}
		$$
		\left({\mathbf{s}_{i+1}^t}\right)^2 = \beta_2 \left( \mathbf{s_{i}^t} \right)^2 + (1 - \beta_2) \mathbf{v_{i+1}^t}, \forall \quad t \in [M]; 
		$$
        \STATE \textbf{Compute the statistics for the counter part $\mathbf{B}_i^t$;}  
		\STATE \textbf{Send the collected statistics to Server.}
}
\end{algorithmic}
\textbf{Server side:}
\begin{algorithmic}[1]
{
    \STATE \textbf{Generate the synthetic gradients:}
    $$
    \hat{\mathbf{A_i^t}} \sim \GN(\mathbf{m_{A,i}^t}, \mathbf{s_{A,i}^t}) \quad \hat{\mathbf{B}}_i^t \sim \GN(\mathbf{m_{B,i}^t}, \mathbf{s_{B,i}^t});
    $$
    \STATE \textbf{Train the AutoEncoder with synthetic gradients:}
    $$
    \min \|  \mathbf{Dec} \circ \mathbf{Enc}(\begin{bmatrix}\hat{\mathbf{A}}_i^t , \hat{\mathbf{B}}_i^t \end{bmatrix}) - \begin{bmatrix} \hat{\mathbf{A}}_i^t ,\hat{\mathbf{B}}_i^t \end{bmatrix} \|;
    $$
    \STATE \textbf{Send $\mathbf{Enc}$ to all clients.}
}
    
\end{algorithmic}
\caption{RandomPrior($[\mathbf{A}_i^t, \mathbf{B}_i^t], \beta_1, \beta_2, h_1, h_2$)}
\label{alg:aux_params}
\end{algorithm}

\begin{algorithm}[t]
\textbf{Input}: 
initial foundation model parameters $\mathbf{W}^0$;
$\mathbf{Enc}$: pretrained encoder with random prior; 
$\mathbf{Dec}$: pretrained decoder with random prior; 
$\sigma$: DP noise multiplier; 
$\eta^t$: learning rate;
$p$: client selection probability ;
$T$: iteration step;  \\
 \textbf{Output} 
Differential private model parameters $\mathbf{w}_i^{T}$. \\
\textbf{Initialize}: 
$\mathbf{W}_i^0 = \mathbf{w}^0$ . \\
\textbf{for communication round $t \in [N]$}:\\
\textbf{Client side}:\\
\textbf{for each client $i \in [M]$}: 
\begin{algorithmic}[1]
{
	\STATE \textbf{Sample a subset of samples $\mathcal{I}_t$ with probability $p$}; 
	\FOR{\textbf{each client $k  \in \mathcal{I}_t$}}
	        	\STATE \textbf{Compute low rank gradient per client:} 
        		$$
        		\mathbf{G_i^t} \approx \mathbf{A_i^t} \mathbf{B_i^t}; 
        		$$
                \STATE \textbf{Compress local gradients:}
                $$
                \mathbf{\dot{A}_i^t} = \mathbf{Enc}(\mathbf{A_i^t}), \quad \mathbf{\dot{B}_i^t} = \mathbf{Enc}(\mathbf{B_i^t})
                $$
         	\STATE \textbf{Clip local gradients with Eq.~\eqref{eqn:clip} and Eq.~\eqref{eqn:clip2};}
                \STATE \textbf{Noise local gradients with Eq.~\eqref{eqn:noise} and Eq.~\eqref{eqn:noise2};}
                \STATE \textbf{Send the local gradients to server;}
	\ENDFOR
}
\end{algorithmic}
\textbf{Server side}:
 \begin{algorithmic}[1]
{			\STATE \textbf{Aggregate per client gradients according to Eq.~\eqref{eqn:aggre};}
        	\STATE \textbf{Decode the compressed gradients with $\mathbf{Dec}$;}
            \STATE \textbf{Do gradient descent with decompressed gradients as Eq.~\eqref{eqn:sgd};}
            \STATE \textbf{Send the updated model parameters back to client.}
}
\end{algorithmic}
\caption{\modelname($\mathbf{W}_0, \sigma, p, T, \mathbf{Enc}, \mathbf{Dec}$)}
\label{alg:feddr}
\end{algorithm}

\subsection{Differentially Private Federated Fine-tuning with LoRA gradients}
First of all, we lay the foundation algorithm for fine-tuning the LLM model with differential LoRA gradients privately. 
Several works consider training LLMs according to differential privacy constraint. 
For example, in ~\cite{charles2024fine}, the author proposes a sample-level differential privacy and a user-level differential privacy method. 
In our work, our goal is to provide differential client-level privacy for the entire FL system. 
In each client at iteration $t$, the gradient is decomposed to a low-rank decomposition for light transmission. 
$
\mathbf{G}_i^t = \mathbf{A}_i^t \mathbf{B}_i^t, \quad \text{where} \quad \mathbf{A}_i^t \in \mathbb{R}^{n \times r}, \mathbf{B}_i^t \in \mathbb{R}^{r \times n}. 
$
We first compress the gradients with the $\mathbf{Enc}$ trained in Algorithm~\ref{alg:aux_params}.
We then clip the gradient.
\begin{eqnarray}
\label{eqn:clip}
\bar{\mathbf{A}}_i^t &= \rm{Clip}(\mathbf{\dot{A}}_i^t) &= \| \mathbf{\dot{A}}_i^t \| \min(1, \frac{C}{\| \mathbf{\dot{A}}_i \|_F}) ; \\
\label{eqn:clip2}
\bar{\mathbf{B}}_i^t &= \rm{Clip}(\mathbf{\dot{B}}_i^t)  &= \| \mathbf{\dot{B}}_i^t \| \min(1, \frac{C}{\| \mathbf{\dot{B}}_i \|_F}) . 
\end{eqnarray}
Next, we exert an random noise on the gradient before transmitted to the global server. 
\begin{eqnarray}
\label{eqn:noise}
\tilde{\mathbf{A}}_i^t & = \bar{\mathbf{A}}_i^t + \mathcal{N}(0, 4 \sigma_A^2/K^2); \\
\label{eqn:noise2}
\tilde{\mathbf{B}}_i^t &= \bar{\mathbf{B}}_i^t + \mathcal{N}(0, 4 \sigma_B^2/K^2). 
\end{eqnarray}
The magnitude of noise exerted on each client's gradient is computed according to the differential privacy mechanism. 
In our design, we use a homogeneous $\sigma$ for $\mathbf{A}$ and $\mathbf{B}$, as 
$$
\sigma_A = \sigma_B = \rm{\textbf{DPAccountant}}(\epsilon, T, p).
$$
The privacy accountant \textbf{DPAccountant} can be chosen utilizing Gaussian Differential Privacy (GDP)~\cite{dong2019gaussian} or Renyi Differential Privacy (RDP)~\cite{mironov2017renyi}. We elaborate on the calculation procedure in Sect.~\ref{sec:privacy}.
After adding noise, the client sends the differential private gradients to the server. 
When the server receives the differential private gradients, it performs the aggregation and is followed by the denoising process. 
\begin{eqnarray}
\label{eqn:aggre}
\tilde{\mathbf{G}}^t = \frac{1}{N} \sum_{i=1}^N \tilde{\mathbf{G}_i^t} = \frac{1}{N} \sum_{i=1}^N \tilde{\mathbf{A}_i^t} \tilde{\mathbf{B}_i^t}. 
\end{eqnarray}
Then the gradient descent is implemented with the aggregated gradients. 
\begin{equation}
\label{eqn:sgd}
\mathbf{W}^{t+1} = \mathbf{W}^t - \eta^t \mathbf{G}^t. 
\end{equation}
After that, the server sends the updated model parameters to each selected client. Then the system runs into the next iteration. 
This loop continues until we run out the privacy budget $\epsilon$ or the system diverges due to the accumulated random noise. 
We describe the procedure formally in Algorithm~\ref{alg:feddr}. 

\section{Theoretical Analysis}

Here, we derive the theoretical analysis to show how to achieve client-level differential privacy based on our proposed methods. 

\subsection{Priliminary}


\begin{definition} \label{def:l2_sens} ($\ell_2$-sensitivity ~\cite{dwork2014algorithmic}). The $\ell_2$-sensitivity of  a function $f: \D \rightarrow \R^d$ is:
$$
\Delta_2 f = \max_{x,y \in \D, \|x - y \| =1} \| f(x) - f(y)\|_2.
$$
\end{definition}

\begin{definition}\label{def:gaussian_mecha} (The Gaussian Mechanism ~\cite{dwork2014algorithmic})
Given a function $f: \D \rightarrow \R^d$ on a data set $\D$, the Gaussian mechanism is defined as:
$$
\M_G(x, f(\cdot), \epsilon) = f(x) + (Y_1, \cdots, Y_k)
$$
where $Y_i$ are $i.i.d.$ random variables drawn from $\mathcal{N}\left( \sigma^2 \Delta_2 f^2 \right)$ and
$\sigma = \frac{2 \ln(1.25/\delta)}{\epsilon}$. 
\end{definition}

\begin{theorem}~\cite{dwork2014algorithmic}
The Gaussian mechanism defined in Definition~\ref{def:gaussian_mecha} preserves $(\epsilon, \delta)-$differential privacy. 
\end{theorem}

\subsection{Privacy of \modelname}
\label{sec:privacy}

First, we show that, with clip value $1$, the sensitivity of the composition of gradient aggregation in Eq.~\eqref{eqn:aggre} is $2/K$.  

\begin{lemma}\label{lem:sens}
With clip value $1$, the sensitivity of gradient aggregation defined in Eq.~\eqref{eqn:aggre} is $2/K$, with $K$ be the number of selected clients.  
\end{lemma}
Then we show with sensitivity equal $1$ and the noise multiplier $\sigma$, the privacy loss for per iteration is  $G_{1/\sigma}$-DP. 

%
%
%
\begin{lemma} [Privacy per iteration] Suppose $\Noise$ with random variable sampled from Gaussian mechanism $\GN(0 , 4\sigma^2/K^2)$. 
Then \modelname~(Algorithm~\ref{alg:feddr}) for per gradient update satisfies  
 $G_{1/\sigma}$-DP,
where 
$
G_{1/\sigma}\left(\alpha \right) = \Phi(\Phi^{-1}(1 - \alpha) - 1/\sigma)
$
and $\Phi$ denotes the standard normal cumulative distribution function. 
\label{lem:pri_per_iter}
\end{lemma}
By applying Lemma~\ref{lem:pri_per_iter} to mechanism defined in Algorithm~\ref{alg:feddr}, we have the privacy analysis of \modelname~for per gradient update is $G_{1/\sigma}-$DP. 
After that, one can prove the privacy analysis of training equipped with \modelname, with subsampling amplification (for SGD or mini-batch SGD) and the central limit theorem of composition over iteration $T$ via GDP~\cite{bu2019deep}. 
According to the Central Limit Theorem (Theorem 5~\cite{bu2019deep}), we have approximated $G_{\mu}$ for the accumulated privacy loss of \modelname. 
Furthermore, we convert the accumulated $\mu-$GDP back to $(\epsilon, \delta)-$DP with the primal-dual result (Corollary 2.13 in~\cite{dong2019gaussian}).  
Then we have the main result of the privacy analysis of \modelname, which preserves $(\epsilon, \delta)-$DP with noise multiplier $2\sigma/K$, number of iterations $T$, and subsampling rate $p$. 

\begin{figure}[t!]
\begin{subfigure}{.475\textwidth}
  \centering
  \includegraphics[width=.98\linewidth]{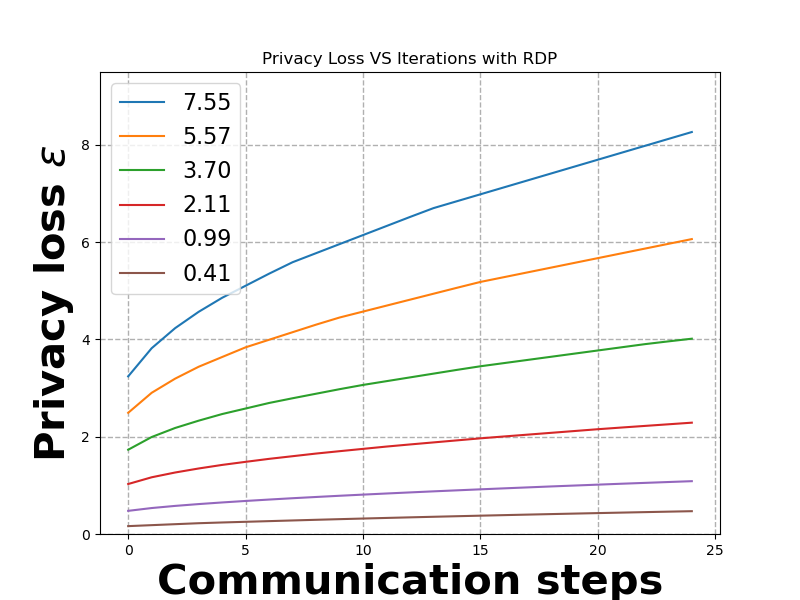}  
\end{subfigure}
\begin{subfigure}{.475\textwidth}
  \centering
  \includegraphics[width=.98\linewidth]{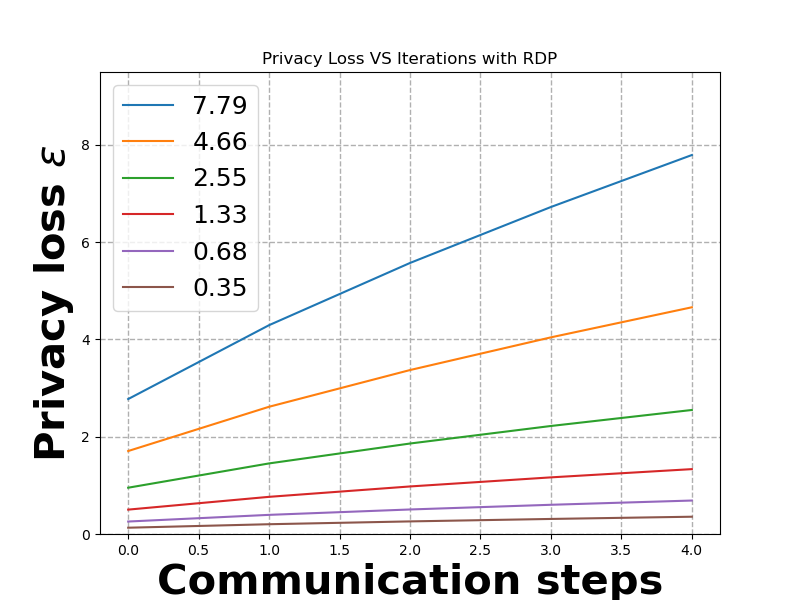}  
\end{subfigure}
\caption{Privacy loss accumulation through RDP. The label denotes the accumulated privacy loss computed by RDP.}
\label{fig:privacy_loss_rdp}
\end{figure}

 \begin{theorem}[Gaussian Differential Privacy~\cite{bu2019deep}]
 \label{thm:privacy_CLT}
 Suppose Algorithm~\ref{alg:feddr} run with number of steps $T$ and Poisson sampling without replacement with probability $p = K/M$, which satisfy $p \sqrt{T} \rightarrow \nu$. Then 
 $C_p \left( G_{1/\sigma} \right)^{\otimes T} \rightarrow G_{\mu}$ uniformly as $T \rightarrow \infty$
 where 
 \begin{align} \label{eqn:mu_sigma}
 \mu = \nu \cdot \sqrt{ (e^{1/\sigma^2} - 1)}. 
 \end{align}
 \end{theorem}
 By solving Eq.~\eqref{eqn:mu_sigma} inversely, we get $\sigma$ via predefined $\epsilon$ and $\delta$. We present the detailed relation between $\sigma$ and $\epsilon, \delta$ in Sect.~\ref{sec:exp_rq2}. Furthermore, we display the accumulation procedure of privacy loss via RDP in Figure~\ref{fig:privacy_loss_rdp}.


\begin{table*}
\centering
\resizebox{1.0\linewidth}{!}{
\begin{tabular}{c ccc ccc }
\hline
Methods & Communication Rounds & Client Selection Fraction & Learning Rates & Low Rank & \# Train Samples & \# Test Samples \\
\hline 
\textbf{LlaMa-Dolly} & 20 & 0.05 & $1.5 \times 10^{-4}$ & 8 & 149 & 13948 \\ 
\textbf{Qwen-MMLU} & 4 & 1.0 & $3\times 10^{-4}$ & 8 & 285 & 14042 \\
\hline 
\end{tabular}}
\caption{Hyperparameters Details of Fine-tuning Examples}
\label{tab:hyperparameters}
\end{table*}

\begin{table*}[t!]
\centering
\begin{tabular}{ccccccc} 
\hline 
Methods & Stem & Social & Humanities & Others & Avgerage & Avg(Hard) \\
\hline 
\textbf{Qwen7B-\modelname-MMLU} & 46.93 & 65.42 & 51.17 & 63.75 & 56.13 &  \\
\textbf{Qwen7B-FedCG-MMLU} & \textbf{47.67}  & \textbf{66.01}  & 50.86 &  64.33 & 56.44 &   \\
\textbf{Qwen7B-LoRA-MMLU} & 47.61 & 65.62 & \textbf{52.09} & 64.4 & \textbf{56.77}  &  \\
\textbf{Qwen7B-Cent-MMLU} & 47.32 & 65.58 & 51.43 & \textbf{64.95} & 56.6 &   \\
\textbf{Qwen7B-Base-MMLU} & 46.62 & 65.52 & 51.2 & 63.69 & 56.07 & \\
\hline 
\textbf{LlaMa7B-\modelname-CEval} & 26.6 & 26.5 & 25.5 & 25.7 & 26.2 & 26.8\\
\textbf{LlaMa7B-FedCG-CEval}   & \textbf{26.8} & 26 & \textbf{26.8} & \textbf{26.5} & \textbf{26.6} & \textbf{26.9} \\
\textbf{LlaMa7B-LoRA-CEval} & 25.9 & \textbf{27.6} & 25.2 & 24.5 & 25.8 & 24.8 \\
\textbf{LlaMa7B-Cent-CEval} & 24.5 & 25.6 & 25.5 & 24.4 & 24.9 & 23.4 \\ 
\textbf{LlaMa7B-Base-CEval}  & 21.6 & 23.4 & 23.9 & 23.3 & 22.8 & 20.3 \\ 
\hline 
\end{tabular}
\caption{Fine-tuning advancements for the Qwen and LlaMa models assessed on MMLU and C-Eval, respectively. The MMLU evaluation comprises five subjects: `Stem', `Social', `Humanities', `Others', and `Average'; whereas the C-Eval includes an additional subject termed `Avg(Hard)'.}
\label{tab:improve}
\end{table*}

\begin{table}
\centering
\begin{tabular}{ccc ccc }
\hline
\multicolumn{3}{c}{\textbf{LlaMa}} & \multicolumn{3}{c}{\textbf{Qwen}} \\
 $\sigma$ &  RDP $\epsilon$ & GDP $\epsilon$ & $\sigma$ &  RDP $\epsilon$ & GDP $\epsilon$ \\ 
\hline 
0.00 & $\infty$ & $\infty$  & 0.00 & $\infty$ & $\infty$ \\ 
0.46 & \textbf{7.55} & 8.00 & 1.10 & \textbf{7.79} & 8.00 \\
0.52 & \textbf{5.57} & 4.00 & 1.63 & 4.66 & \textbf{4.00} \\
0.60 & 3.71 & \textbf{2.00} & 2.64 & 2.55 & \textbf{2.00} \\
0.75 & 2.12 & \textbf{1.00} & 4.47 & 1.33 & \textbf{1.00} \\
1.00 & 1.00 & \textbf{0.50} & 7.70 & 0.69 & \textbf{0.50} \\
1.45 & 0.42 & \textbf{0.25} & 13.24 & 0.36 & \textbf{0.25} \\
\hline 
\end{tabular}
\caption{Comparison of noise multiplier $\sigma$ and privacy budget $\epsilon$ as calculated by RDP and GDP.}
\label{tab:rdp_gdp_compare}
\end{table}

\begin{table*}[t!]
\centering
\begin{tabular}{c|ccccccc} 
\hline 
& Methods & Stem & Social Sciences & Humanities & Others & Average & Avg(hard) \\
\hline 
\multicolumn{7}{c}{\textbf{MMLU}} \\
\hline 
\multirow{6}{*}{1} & 
\textbf{Qwen-LoRA-$\epsilon$-0.25} & 26.2 & 23.24 & 26.8 & 23.82 & 25.22 & \\
& \textbf{Qwen-LoRA-$\epsilon$-0.5} & 25.75 & 23.23 & 26.8 & 24.23 & 25.21 &  \\
& \textbf{Qwen-LoRA-$\epsilon$-1.0} & 25.75 & 23.01 & 26.48 & 24.39 & 25.09 & \\
& \textbf{Qwen-LoRA-$\epsilon$-2.0} & 24.99 & 22.94 & 26.56 & 25.61 & 25.21 &  \\
& \textbf{Qwen-LoRA-$\epsilon$-4.0} & 25.11 & 23.91 & 25.86 & 26.93 & 25.5 & \\ 
& \textbf{Qwen-LoRA-$\epsilon$-8.0} & 25.25 & 24.11 & 26.23 & 26.81 & 25.67 & \\
\hline 
\multirow{6}{*}{2} & 
\textbf{Qwen-FedCG-$\epsilon$-0.25} & 46.62	 & 64.8 & \textbf{52.64} & 63.56  & 56.37 & \\
& \textbf{Qwen-FedCG-$\epsilon$-0.5} & 47.06	& 65.48	 & 52.17 & 64.11 & 56.58 & \\
& \textbf{Qwen-FedCG-$\epsilon$-1.0} & 45.92	& 64.93	& 52.53 & 63.63	 & 56.22 &  \\
& \textbf{Qwen-FedCG-$\epsilon$-2.0} & 47.25	 & 65.87 & 51.75 & 64.04 & 56.55 &  \\
& \textbf{Qwen-FedCG-$\epsilon$-4.0} & 47.35	& 65.42	 & 51.11 & 63.59 & 56.16 & \\ 
& \textbf{Qwen-FedCG-$\epsilon$-8.0} &46.97	& 65.42	 & 51.56 & 63.98 & 56.31 &  \\
\hline 
\multirow{6}{*}{3} & 
\textbf{Qwen-FedDR-$\epsilon$-0.25}	& 40.84 & 59.4 & 47.65 & 59.09 & 51.23 & \\
& \textbf{Qwen-FedDR-$\epsilon$-0.5} &  42.08	& 60.09	& 47.01 & 59.28	& 51.48 &  \\
& \textbf{Qwen-FedDR-$\epsilon$-1.0} & 44.02 & 61.48 & 49.58 & 61.08 & 53.48 &  \\
& \textbf{Qwen-FedDR-$\epsilon$-2.0} & 46.14 & 63.89 & 50.86 & 62.43 & 55.22 &  \\
& \textbf{Qwen-FedDR-$\epsilon$-4.0} & 47.28 & 65.25 & 51.07 & 64.04 & 56.2  & \\
& \textbf{Qwen-FedDR-$\epsilon$-8.0} & 47.35 & \textbf{66.23} & 51.39 & 63.82 & 56.48  & \\
\hline 
\multirow{5}{*}{4} & 
\textbf{Qwen-\modelname-$\epsilon$-$\infty$} & 46.93 & 65.42 & 51.17 & 63.75 & 56.13 & \\
& \textbf{Qwen-FedCG-$\epsilon$-$\infty$} & \textbf{47.67}  & 66.01  & 50.86 &  64.33 & 56.44 &  \\
& \textbf{Qwen-LoRA-$\epsilon$-$\infty$} & 47.61 & 65.62 & 52.09 & 64.4 & \textbf{56.77} &  \\
& \textbf{Qwen-Cent-$\epsilon$-$\infty$} & 47.32 & 65.58 & 51.43 & \textbf{64.95} & 56.6  & \\
& \textbf{Qwen-Base} & 46.62 & 65.52 & 51.2 & 63.69 & 56.07 & \\
\hline
\multicolumn{7}{c}{\textbf{C-Eval}} \\
\hline 
\multirow{2}{*}{5} & 
\textbf{LlaMa-FedLoRA-$\epsilon$-4.0} & 24.3 & 24.2 & 24.4 & 23.9 & 24.2 & 23.8 \\
& \textbf{LlaMa-FedLoRA-$\epsilon$-8.0} & 24.5 & 25.7 & 26.1 & 25.9 & 25.4 & 24 \\
\hline 
\multirow{4}{*}{6} & 
\textbf{LlaMa-FedCG-$\epsilon$-0.25} & \textbf{26.8} & 25.3 & \textbf{26.4} & \textbf{26.6} & \textbf{26.4} & \textbf{27.2} \\
& \textbf{LlaMa-FedCG-$\epsilon$-0.5} & 26.6 & 26.5 & 25.5 & 25.7 & 26.2 & 26.8 \\
& \textbf{LlaMa-FedCG-$\epsilon$-1.0} & 24.9 & 24.8 & 24.3 & 24.5 & 24.7 & 25.8 \\
& \textbf{LlaMa-FedCG-$\epsilon$-8.0} & 21.7 & 23.3 & 23.8 & 23.3 & 22.8 & 20.3 \\ 
\hline 
\multirow{4}{*}{7} & 
\textbf{LlaMa-\modelname-$\epsilon$-0.25} & 	26.6 & 26.5 & 25.4 & 25.7 & 26.1 & 26.8 \\
& \textbf{LlaMa-\modelname-$\epsilon$-0.5} & 22.2 & 24 & 24.4 & 23.9 & 23.3 & 20.3 \\
& \textbf{LlaMa-\modelname-$\epsilon$-2.0} & 24.7 & 25.3 & 24.8 &	23.9 & 24.7 & 25 \\
& \textbf{LlaMa-\modelname-$\epsilon$-8.0} & 26.6 & 26.2 & 25.6 & 25.8 & 26.1 & 26.8 \\
\hline 
\multirow{5}{*}{8} & 
\textbf{LlaMa-\modelname-$\epsilon$-$\infty$} & 26.6& 26.5 & 25.5 & 25.7 & 26.2	& 26.8  \\
& \textbf{LlaMa-FedCG-$\epsilon$-$\infty$} &26.6 & 26.5 & 25.5 & 25.7 & 26.2 & 26.8 \\
& \textbf{LlaMa-LoRA-$\epsilon$-$\infty$}& 25.9 & \textbf{27.6} & 25.2 & 24.5 & 25.8 & 24.8 \\
& \textbf{LlaMa-Cent-$\epsilon$-$\infty$} & 24.5 & 25.6 & 25.5 & 24.4 & 24.9 & 23.4 \\ 
& \textbf{LlaMa-Base} & 21.6 & 23.4 & 23.9 & 23.3 & 22.8 & 20.3 \\ 
\hline  
\end{tabular}
\caption{Tests across Various Privacy Budget Levels, assessed through MMLU (Rows 1 to 4) and C-Eval (Rows 5 to 8). The MMLU assessment includes 5 subjects, whereas the C-Eval includes an additional one called 'Avg(hard)'.}
\label{tab:dp}
\end{table*}

\section{Experiments}

In this segment, we perform extensive experiments to address the following research inquiries.

\begin{itemize}

\item{\textbf{RQ1}} How does the \modelname~train by Random Prior work in the fine-tuning scenario? 
\item{\textbf{RQ2}} How much noise do we exert on the gradient to guarantee DP with various budgets? 
\item{\textbf{RQ3}} How does the random mechanism to achieve DP influence the FedLLM fine-tuning? 
\item{\textbf{RQ4}} What is the difference in performance between the AutoEncoder trained with the informative prior and with the random prior? 
\item{\textbf{RQ5}} What is the gain in communication efficiency of the proposed method? 
\end{itemize}

\subsection{Experimental Settings}
Now we present the details of experimental settings including foundation models, datasets, evaluation benchmarks, and hyper-parameters that we utilized throughout all papers. 

For the foundational models, we adopt the fine-tuning strategy based on LlaMa-7B and Qwen-7B. 
LlaMa is an open and efficient LLM foundation model developed by Meta with parameter sizes ranging from 7B to 65B~\cite{touvron2023llama}. 
Qwen is developed and released by Alibaba Group~\cite{bai2023qwen}.
Due to the limited computational resource, we carry out the experiments on the 7B version. 
For the fine-tuning of our large language models (LLMs), we leveraged two primary datasets: MMLU-train \cite{hendrycks2020measuring} and Databricks-dolly-15k \cite{conover2023free}. The MMLU-train dataset encompasses 57 tasks in diverse disciplines, such as elementary mathematics, United States history, computer science, legal studies, and more. Meanwhile, Databricks-dolly-15k is structured into eight categories, including brainstorming, classification, closed QA, creative writing, general QA, information extraction, open QA, and summarization. To evaluate the performance of our LLM, we used the C-Eval \cite{zhang2023building} and MMLU \cite{hendrycks2020measuring} benchmarks, which provide comprehensive, multidisciplinary assessments designed specifically to evaluate LLM capabilities.
Drawing on established research, we select the hyperparameters for FedLLM~\cite{zhang2023building, Shepherdgithub}. Details on the fine-tuning process and the hyperparameters chosen are presented in Table~\ref{tab:hyperparameters}. 
The compared methods are `\textbf{Cent}', which stands for centralized fine-tuning, `\textbf{LoRA}', representing subspace decomposition of local gradient parameters, `\textbf{FedCG}', indicating gradient compression with a knowledgeable prior developed in~\cite{wu2024cgfedllmcompressgradientsfederated}, and `\modelname', referring to our approach involving gradient compression with random prior.
Each experiment is labeled in the format `\textbf{A-B-C}', where \textbf{A} represents the base model, \textbf{B} signifies the tuning technique, and \textbf{C} indicates the evaluation criteria.

\begin{figure*}[t!]
\begin{subfigure}{.245\textwidth}
  \centering
  \includegraphics[width=.98\linewidth]{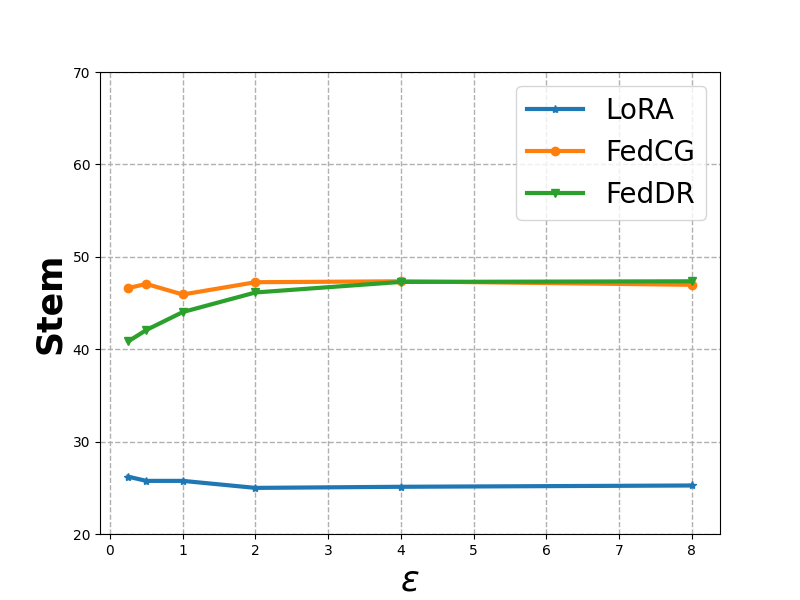}  
  \caption{Stem}
\end{subfigure}
\begin{subfigure}{.245\textwidth}
  \centering
  \includegraphics[width=.98\linewidth]{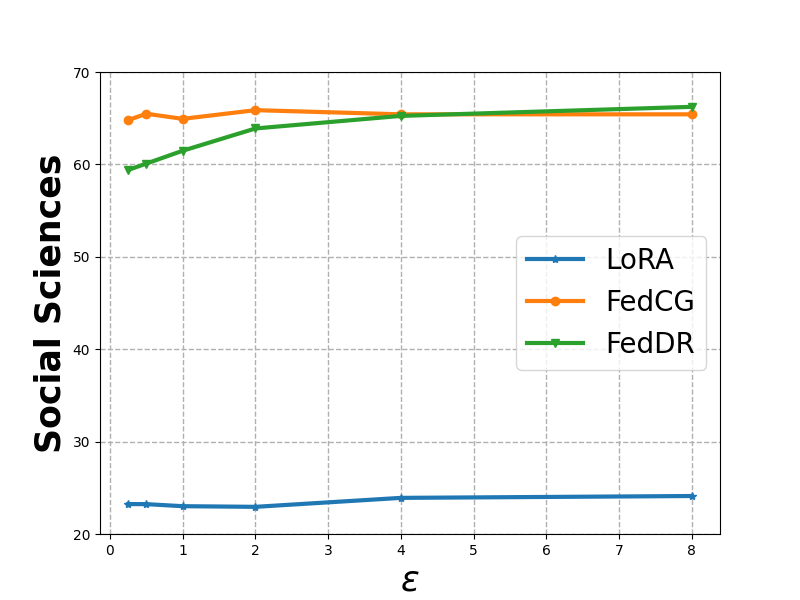}  
  \caption{Social Sciences}
\end{subfigure}
\begin{subfigure}{.245\textwidth}
  \centering
  \includegraphics[width=.98\linewidth]{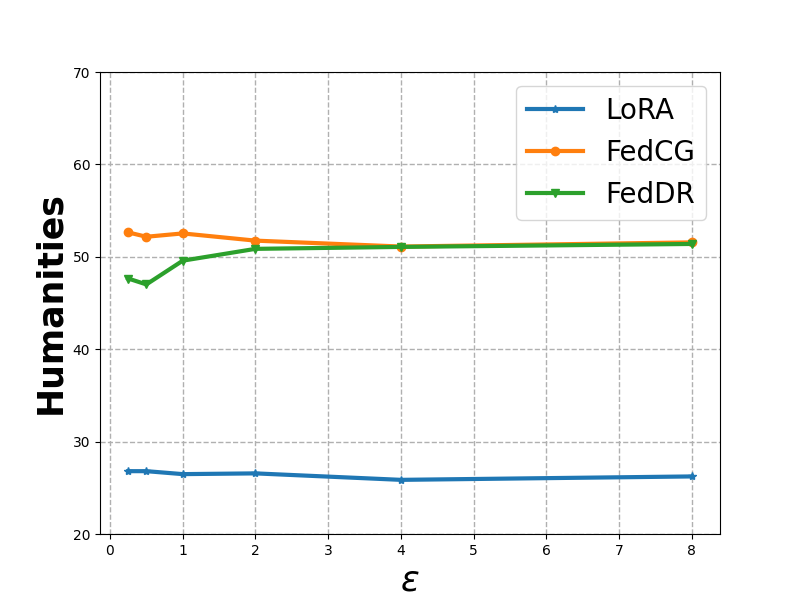}  
  \caption{Humanities}
\end{subfigure}
\begin{subfigure}{.245\textwidth}
  \centering
  \includegraphics[width=.98\linewidth]{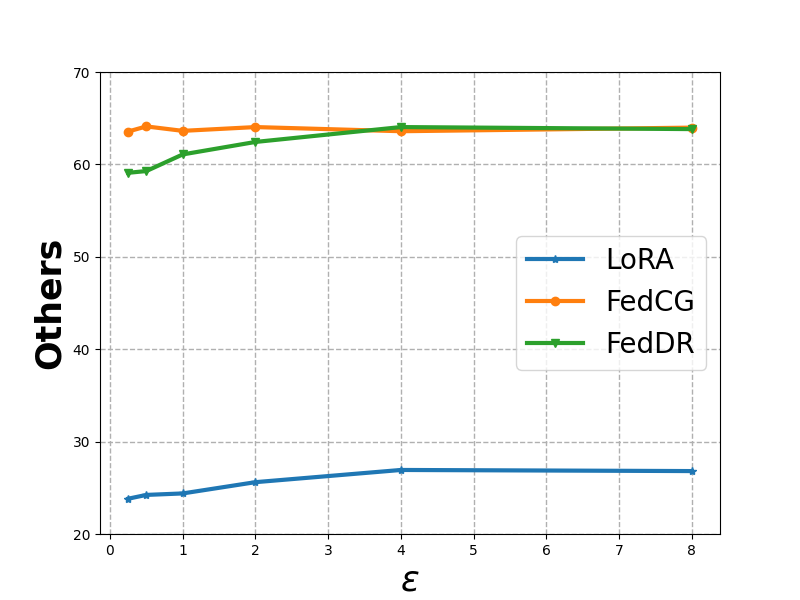}  
  \caption{Others}
\end{subfigure}
\caption{MMLU Performance Analysis across four disciplines. The y-axis shows the evaluation score, while the x-axis shows the privacy budget. The yellow line denotes \textbf{FedCG}, the green line denotes \modelname, and the blue line denotes \textbf{LoRA}.}
\label{fig:compare_mmlu}
\end{figure*}

\subsection{Foundation Models Improvement (RQ1)}
\label{sec:exp_rq2}
In this section, we demonstrate how various fine-tuning strategies enhance model performance over the foundational ones. We detail the fine-tuning of the Qwen model with MMLU datasets and its evaluation in MMLU, as well as the fine-tuning with dolly-15k datasets and evaluation on C-Eval, in Table~\ref{tab:improve}. From Table~\ref{tab:improve}, it is evident that fine-tuning methods significantly improve accuracy compared to foundational methods. For instance, in the second part (C-Eval) of Table~\ref{tab:improve}, the accuracy improvement of \textbf{FedCG} is 3.8 for the `Average' subject and 6.3 for `Avg(Hard)' relative to the foundational model. In the first part (MMLU) of Table~\ref{tab:improve}, the \textbf{LoRA} method exhibits an accuracy increment of 0.7 in the `Average' subject while \textbf{FedCG} shows an increment of 0.37 compared to the \textbf{Qwen7B-Base-MMLU} model. Moreover, the \modelname~with an autoencoder trained using random prior information demonstrates a comparable fine-tuning performance to \textbf{FedCG} with an autoencoder trained on raw gradients, outperforming \textbf{LoRA} without gradient compression. For example, in the C-Eval evaluation, the highest score of 26.6 for the `Average' subject is achieved by \textbf{FedCG}, while the second highest score of 26.2 is achieved by \modelname, which surpasses \textbf{LoRA}'s performance of 25.8. A similar trend is observed for the `Avg(Hard)' subject.

\subsection{Privacy Accounting via RDP or GDP (RQ2)}
This section discusses how to account for privacy loss in the FedLLM system. We employ the RDP~\cite{mironov2017renyi} and GDP~\cite{dong2019gaussian} methods to account for loss of privacy. We provide a table showing the one-to-one relationship between the privacy budget $\epsilon$ and the noise multiplier for both RDP and GDP (Table~\ref{tab:rdp_gdp_compare}). The calculation of GDP follows the steps described in Sect.~\ref{sec:privacy}. Initially, we set the privacy budget from the set $\{ 2^k | k \in \mathcal{N}, -3 \leq k \leq 2 \}$ (columns 2 and 5). Next, we determine the magnitude of the noise multiplier for each client denoted as $\sigma$. Each client has an evenly distributed amount of Gaussian noise added. After obtaining the added noise (columns 1 and 4), we use RDP to aggregate the privacy cost measured in RDP according to established research~\cite{mironov2017renyi,balle2018improving} and derive the RDP $\epsilon$ (columns 3 and 6). From Table~\ref{tab:rdp_gdp_compare}, we note that when $\sigma$ is small, RDP indicates a lower loss of privacy than GDP. In contrast, GDP results in a lower privacy loss when $\sigma$ is large and the privacy budget is small. In subsequent sections, we use $\epsilon$ derived via GDP. It should also be noted that RDP demonstrates a lesser privacy loss when $\epsilon$ is large. Furthermore, we illustrate the privacy loss accumulation process in Figure~\ref{fig:privacy_loss_rdp}. 

\subsection{Overall Influence of Differential Privacy (RQ3)}

In Table~\ref{tab:dp}, we present the training results with various privacy budgets of the set $\{0.25, 0.5, 1.0, 2.0, 4.0, 8.0\}$. Smaller privacy budgets ensure stronger privacy guarantees. 
Based on the results shown in Table~\ref{tab:dp}, we observe that the inclusion of Gaussian noise in the low-rank decomposition of gradients to provide differential privacy leads to a decline in model performance. As illustrated in the first row of Table~\ref{tab:dp}, adding Gaussian noise directly to the LoRA gradients results in the `Average' score dropping from 56.77 to 25.67. This shows that Gaussian noise has a negative impact on model training. 
For the LlaMa model, there are no evaluation results for LoRA methods with small $\epsilon$ due to the substantial noise applied to the gradients.

\subsection{Comparison Between \textbf{FedCG} and \modelname~(RQ4)}

With the use of an informative pre-trained AutoEncoder, \textbf{FedCG} achieves performance comparable to the nonprivate scenario. For example, the scores obtained by \textbf{Qwen-FedCG-$\epsilon$-0.5} are close to those of \textbf{Qwen-LoRA-$\epsilon$-$\infty$}. This is attributed to the strong denoising capability of the informative AutoEncoder. A similar pattern can be observed for the LlaMa example. 
Moreover, \textbf{FedCG} displays consistent performance across different privacy budgets. This consistency is likely a result of the denoising capabilities of the informative AutoEncoder. 
In contrast to \textbf{FedCG}, \modelname~exhibits a downward trend in evaluation scores as privacy budgets become more stringent. This aligns with the general understanding that a smaller privacy budget results in more significant Gaussian noise being added, thereby impairing overall model performance.

\subsection{\modelname~with Large Privacy Budget (RQ4)}
Most notably, even when employing an AutoEncoder trained with a Random Prior, \modelname~efficiently alleviates the performance degradation caused by differential privacy for moderate privacy budgets of $\epsilon = 1.0$ and $2.0$. Interestingly, with looser privacy budgets of $\epsilon = 4.0$ and $8.0$, \modelname~surpasses the non-privacy benchmarks, such as \textbf{Compress-$\epsilon$-$\infty$} and \textbf{Cent-$\epsilon$-$\infty$}. This happens because looser privacy budgets lead to a smaller amount of random noise, thus improving the generalizability of the fine-tuned model.

\subsection{Communication Efficiency Gain (RQ5)}
In this section, we demonstrate the communication efficiency gain (CEG) of our method \modelname~in comparison to \textbf{FedCG}. In the initial stage, which involves collecting gradients and transmitting them to the server for autoencoder pre-training, for the LlaMa model with low-rank parameters $r=8$, \textbf{FedCG} requires $4096 \times 8 \times 2 \times 4 \times 32  \times 20 = 167,772,160 \approx 160 MB$ of parameters. In contrast, for \modelname, we only send the mean and standard deviation for each layer and the gradients of each epoch, with a total parameter count of $2 \times 2 \times 4 \times 32 \times 20 = 10,240 \approx 0.078 MB$. We reduce the communication complexity in the AutoEncoder preprocessing stage to $6.10 \times 10^{-5}$, illustrating the substantial efficiency gain of our proposed AutoEncoder training with Random Prior. We summarize this comparison of model parameters, storage, and efficiency gain in Table~\ref{tab:computational_efficiency}.

\begin{table}[t!]
\centering
\begin{tabular}{c c c } 
\hline 
 & \#Parameters & Storage \\
 \hline 
 \textbf{FedCG} & 167,772,160 & 160 MB \\
 \modelname & 10,240  &  0.078 MB \\
 \textbf{CEG} & \multicolumn{2}{c}{$6.10 \times 10^{-5}$} \\
 \hline 
\end{tabular}
\caption{Efficiency Improvement of \modelname~During the AutoEncoder Pretraining Phase.}
\label{tab:computational_efficiency}
\end{table}



\section{Conclusion}

In this paper, we introduce a practical learning-based gradient compression method to achieve client-level differential privacy during fine-tuning of LLMs. Our approach offers high privacy guarantees, minimal communication overhead, and manageable computational demands. We utilize statistical data of the local gradients to train an AutoEncoder, preventing reverse attacks from the AutoEncoder on raw gradients as well as from raw gradients to the original input data. In addition, we apply Gaussian noise to compressed gradients to ensure user-level differential privacy. We quantify privacy loss using GDP and RDP, offering a thorough privacy analysis. We propose a one-stage fine-tuning algorithm by integrating AutoEncoder training with LLM fine-tuning into a cohesive system. Moreover, our method can be extended to the Large Vision Model (LVM) and Large Multimodal Model (LMM) in the future.

\section{Acknowledgments}
The work is supported by the Zhejiang Province Key Research and Development Plan (No. 2024SSYS0010).

\bibliographystyle{plain}
\bibliography{aaai25}
\newpage

\section{Supplementary Material}

\begin{figure*}[t!]
     \centering
     \includegraphics[width=\textwidth]{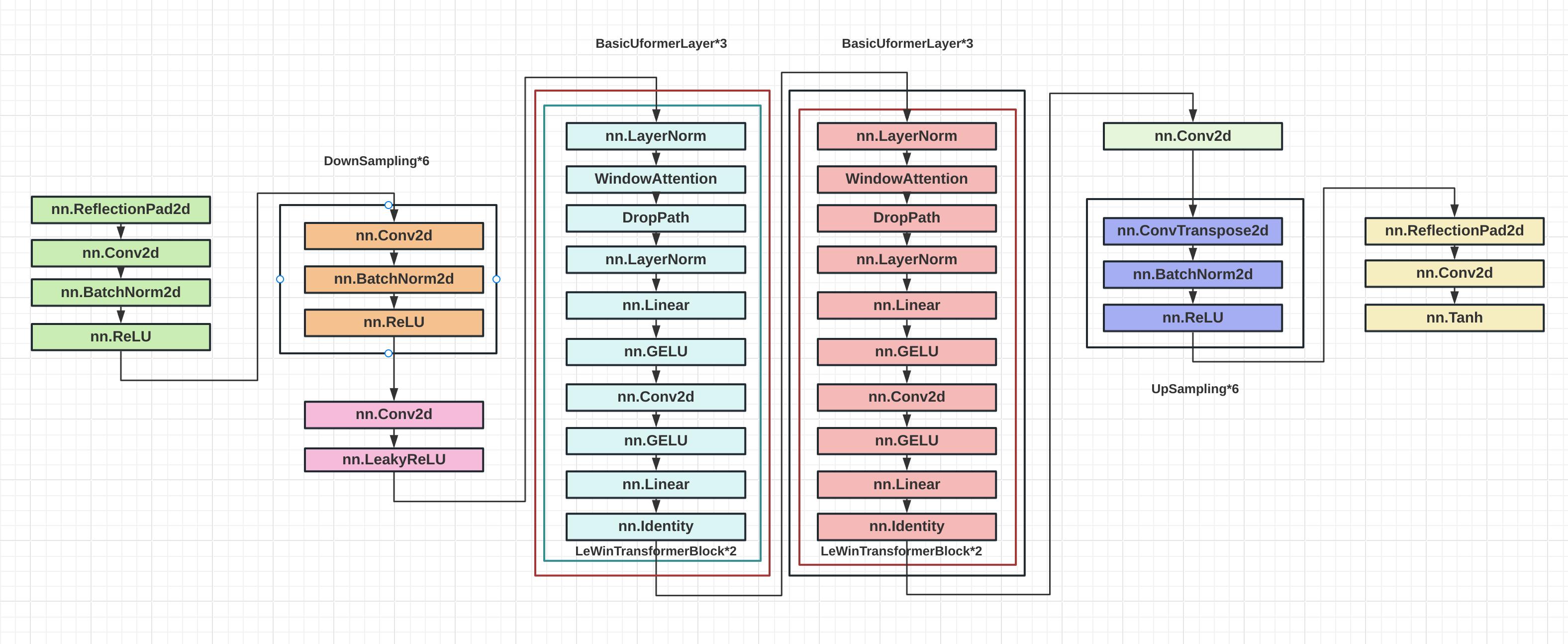}
     \caption{Architecture of AutoEncoder with Uformer~\cite{wang2022uformer}}
     \label{fig:archi_ae_uformer}
\end{figure*}

\subsection{Training Details}
\label{sec:append_train}

In this part, we detail the training procedures of the experiments discussed in the main document.

\subsubsection{Hyper-parameters}

The hyperparameters for fine tuning are detailed in Table~\ref{tab:append_hyper}. Using the FedLLM hyperparameters, the noise multiplier $\sigma$ for the privacy hyperparameters is calculated via GDP~\cite{dong2019gaussian}, as shown in the main content, Table 3.
In industry DP applications, the standard privacy budgets used include $\epsilon =4.0$ for Apple Emoji, $\epsilon = 2.0$ for Apple Health, and $\epsilon = \ln 3$ for Google's RAPPOR~\cite{erlingsson2014rappor}. However, in our experiments, we also tested significantly tighter values of $\epsilon$ across all data sets as $\{2^{-k}, k \in \mathcal{N}, -3 \leq k \leq 2 \}$ to validate the robustness of our method.

\begin{table*}[t!]
\centering
\resizebox{1.0\linewidth}{!}{
\begin{tabular}{c   c c c c c c c}   
\toprule
& Communication Rounds  &  Client Selection fraction  &  Local Batch Size  & Local Micro Batch Size  &  Local Learning Rate  &  Low Rank Parameters  &  Local Number Epochs   \\\midrule 
\textbf{LlaMa-Dolly}  & 20 & 0.05 & 32 & 16 & $1.5 \cdot 10^{-4}$ & 8 & 1 \\
\textbf{Qwen-MMLU}  & 8 & 1.0 & 8 & 4 & $3. \cdot 10^{-4}$  & 8 & 2 \\
\bottomrule
\end{tabular}
}
\caption{Hyper-parameters in FedLLM}
\label{tab:append_hyper}
\end{table*}

\subsubsection{Computation Power}

One of our servers is equipped with an NVIDIA GPU (serial number 1) with 24 GB of memory. This allows us to perform parallel processing for tasks that require high computational power, such as pre-training the AutoEncoder in stage one and federated fine-tuning in stage two.
Our computational infrastructure runs on Ubuntu Linux 22.04. Additionally, to optimize the performance of specific machine learning components in our simulations, we have implemented PyTorch 2.2.1 with CUDA 12.4 to seamlessly integrate GPU processing into our workflows.

\subsubsection{Training and Fine-tuning Data}

We utilize the identical dataset for both AutoEncoder pre-training and FedLLM fine-tuning.
The data set applied with the LlaMa model is Dolly-bricks-15k, whereas the data set for the Qwen model is the MMLU data set.
The Databricks-dolly-15k~\cite{conover2023free} dataset encompasses eight distinct categories: brainstorming, classification, closed QA, creative writing, general QA, information extraction, open QA and summarization. This data set is segregated into 100 segments using the widely used category Dirichlet allocation technique~\cite{wang2020federated}.
The MMLU dataset~\cite{hendrycks2020measuring} spans 57 tasks, including elementary mathematics, US history, computer science, law, and others. This data set is randomly partitioned into 3 segments.

\begin{table*}[t!]
\centering
\begin{tabular}{c c c c c c c}  
\toprule
 & Layer & In Channels & Out Channels  & Kernel Size & Stride  \\ \midrule
\multirow{8}{*}{Encoder}  
& \multirow{7}{*}{DownSampling Layers} 
& conv2d & 1 & 1 & (3,3) & (1,1) \\ 
& & conv2d & 1 & 2 & (3,3) & (2,2) \\ 
& & conv2d & 2 & 4 & (3,3) & (2,2) \\ 
& & conv2d & 4 & 8 & (3,3) & (2,2) \\ 
& & conv2d & 8 & 16 & (3,3) & (2,2) \\ 
& & conv2d & 16 & 32 & (3,3) & (2,2) \\   
& & conv2d & 32 & 64 & (3,3) & (2,2) \\
&  \multicolumn{6}{c}{Uformer Block $\times 3$} \\
\midrule
\midrule
\multirow{8}{*}{Decoder} 
&  \multicolumn{6}{c}{Uformer Block $\times 3$} \\
& \multirow{7}{*}{UpSampling Layers} 
& deconv1 & 4 & 32 & (3,3) & (2,2) \\ 
& & deconv2d & 64 & 32 & (3,3) & (2,2) \\ 
& & deconv2d & 32 & 16 & (3,3) & (2,2) \\  
& & deconv2d & 16 & 8 & (3,3) & (2,2) \\  
& & deconv2d & 8 & 4 & (3,3) & (2,2) \\ 
& & deconv2d & 4 & 2 & (3,3) & (2,2) \\  
& & deconv2d & 2 & 1 & (3,3) & (2,2) \\ 
& & conv2d & 1 & 1 & (7,7) & (1,1) \\
\bottomrule
\end{tabular}
\caption{AutoEncoder with Uformer~\cite{wang2022uformer}}
\label{tab:archi-uformer}
\end{table*}

\subsection{Privacy Analysis Details}
\label{sec:append_privacy}

This section provides an in-depth analysis of privacy loss using GDP~\cite{dong2019gaussian}.

The random mechanism depicted in the main context, Algorithm 2 can be abstracted in the following operator form. 

\begin{equation}
\label{eqn:random_mech}
\mathcal{M} :=  \Dec \circ \Aggre \circ \Noise \circ  \Clip  \circ \Enc .
\end{equation}

$\Enc$ represents the compression phase through the encoder. $\Clip$ signifies the clipping stage. $\Noise$ incorporates random Gaussian noise into the gradients. $\Aggre$ indicates the aggregation step by summing all the gradients from the chosen clients. $\Dec$ signifies the decompression phase with the decoder. Randomness is present in the $\Noise$ procedure, while all other steps are deterministic.

\subsubsection{Proof of Lemma 4.1}

\begin{proof}
Suppose $\mathcal{D}'$ is a data set adjacent to $\mathcal{D}$ with one date changing, which leads to
only one $\mathbf{G}_i^t$ change. 
By clipping with the value $1$, the clipped low rank parameters have a Frobenious norm less than or equal to 1.
$$
\| \mathbf{\bar{G}_i^t} \|_F = \| \mathbf{\bar{A}_i^t} \mathbf{\bar{B}_i^t} \|_F \leq \| \mathbf{\bar{A}_i^t} \|_F \| \mathbf{\bar{B}_i^t} \|_F \leq 1.
$$
$$
\| \sum_{i=1}^K \mathbf{\bar{G}^t_{\mathcal{D}}} - \sum_{i=1}^K \mathbf{\bar{G}^t_{\mathcal{D'}}} \|_F \leq 2, 
$$
since $\| \mathbf{\bar{G}_i^t} \|_F  \leq 1$ after clipping.  
Thus, the sensitivity of the gradient aggregation in \modelname~defined in Eq.\eqref{eqn:random_mech} is $2/K$, where $K$ is the number of selected clients. 
\end{proof}

\subsubsection{Proof of Lemma 4.2}
\label{proof:lem_3}

 \begin{proof}
 By Theorem 2.7 in GDP~\cite{dong2019gaussian}, with the addition of DP noise $4 \sigma^2 /K^2$, the gradient aggregation step is $G_{1/\sigma}$-DP. 
 Furthermore, the decoder $\mathbf{Dec}$ is deterministic after the gradient aggregation step. 
 By the post-process property (Proposition 2.8~\cite{dong2019gaussian}), we have  that 
 \modelname~for per gradient update satisfies $G_{1/\sigma}$-DP.
 \end{proof}

\subsubsection{Central Limit Theorem for GDP}

 \begin{theorem}[Central Limit Theorem for GDP~\cite{bu2019deep}]
 \label{thm:privacy_CLT_1}
 Suppose Algorithm~1 run with number of steps $T$ and Poisson sampling without replacement with probability $p = K/M$, which satisfy $p \sqrt{T} \rightarrow \nu$. Then 
 $C_p \left( G_{1/\sigma} \right)^{\otimes T} \rightarrow G_{\mu}$ uniformly as $T \rightarrow \infty$
 where 
 \begin{align} \label{eqn:mu_sigma_1}
 \mu = \nu \cdot \sqrt{ (e^{1/\sigma^2} - 1)}. 
 \end{align}
 \end{theorem}

 \subsubsection{Proof of Theorem Theorem 4.2}
\label{proof:thm_1}
 
 \begin{proof}
 By Lemma 4.1 and Lemma 4.2, we find that the update per gradient for \modelname~is $G_{1/\sigma}$-DP.
 According to the GDP subsampling and composition theorem~\cite{dong2019gaussian}, with the sampling rate $p$ and iteration $T$, the overall loss of privacy is $C_p \left(G_{1/\sigma}\right)^{\otimes T}$. 
 By the Central Limit Theorem (Theorem 5~in \cite{bu2019deep}), if $p \sqrt{T} \rightarrow v, and~   \mu = \nu \cdot \sqrt{ (e^{1/\sigma^2} - 1)},$ 
   $C_p \left( G_{1/\sigma} \right)^{\otimes T} \rightarrow G_{\mu}$ as $T \rightarrow \infty$. 
  \end{proof}

\subsubsection{Choice of Privacy Accountant}

By GDP~\cite{bu2019deep}, one can compute the noise multiplier $\sigma$ by the sampling rate $p$, the number of rounds $T$, the privacy budget and $\delta$.
For comparison, we present the $\epsilon$ accumulated with RDP~\cite{mironov2017renyi} under the same noise magnitude $\sigma$ in the main context, Table 3, which shows a looser accumulation compared to GDP.
Small $\epsilon$ and $\delta$ provide a strong privacy guarantee but can degrade the utility of the model. 
The frequent privacy budget used in academia and industry is $(\epsilon, 10^{-5})$ where $\epsilon$ is $4.0$ for Apple Emoji, $2.0$ for Apple Health~\cite{tang2017privacy}, and $\ln3$ for Google RAPPOR~\cite{erlingsson2014rappor}.
In our experiments, we achieve a stronger privacy guarantee with $\epsilon$ as small as $ \{ 0.25, 0.5 \}. $

\subsubsection{Client-level Differential Privacy}

To conclude, we demonstrate that the random mechanism described by Eq.~\ref{eqn:random_mech} ensures $\{\epsilon, \delta\}-$DP by utilizing the Gaussian mechanism as specified in Definition 4.2 of the main text as the $\Noise$ operator, while all other operators are deterministic. Additionally, the privacy focus in our discussion is per client rather than per sample because the $\Aggre$ operator functions during the server-side aggregation of all client gradients. From the above analysis, we conclude the privacy proof for our proposed methods, which comply with differential privacy at the client level.

\subsection{Architecture of Encoder and Decoder}
\label{sec:append_archi}

We refine the AutoEncoder using the Uformer architecture defined in~\cite{wang2022uformer}. The architectural details are shown in Table~\ref{tab:archi-uformer}, and the training methodology is illustrated in Figure~\ref{fig:archi_ae_uformer}. This iteration of the AutoEncoder preserves the upsampling and downsampling layers inherent to the ResNet-based AutoEncoder while replacing the core components with Basic Uformer units, each comprising two LeWin Transformer Blocks. The encoder integrates the DownSampling layers with three Basic Uformer blocks, whereas the decoder incorporates three Basic Uformer blocks in conjunction with the UpSampling layers. 
Moreover, we test on different architectures by changing the number of Uforme blocks or the number of Upsampling and Downsampling and choose the best one with 7 Downsampling + 6 Uformer + 7 Upsampling by balancing efficiency and efficacy.

\subsection{Gradient Distribution Visualization}

\begin{figure*}[htbp]
     \begin{subfigure}[b]{0.24\textwidth}
         \centering
         \includegraphics[width=\textwidth]{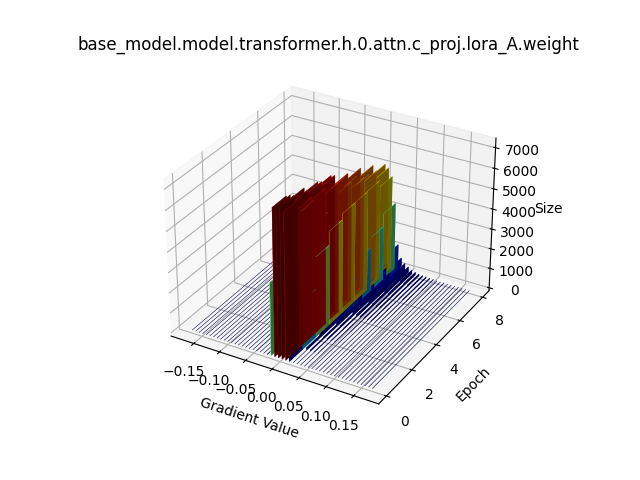}
     \end{subfigure}
     \begin{subfigure}[b]{0.24\textwidth}
         \centering
         \includegraphics[width=\textwidth]{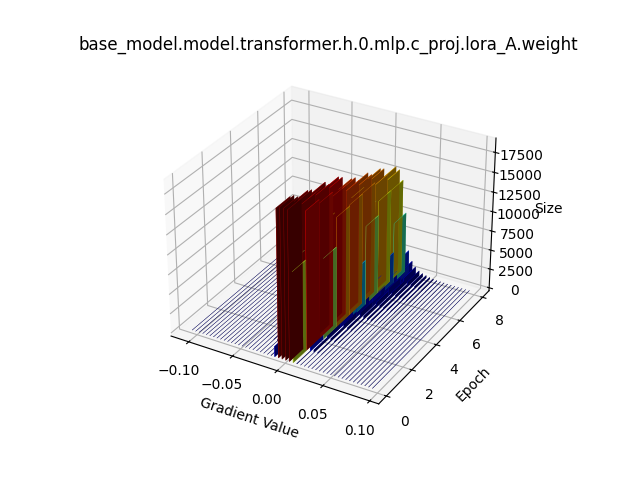}
     \end{subfigure}
     \begin{subfigure}[b]{0.24\textwidth}
         \centering
         \includegraphics[width=\textwidth]{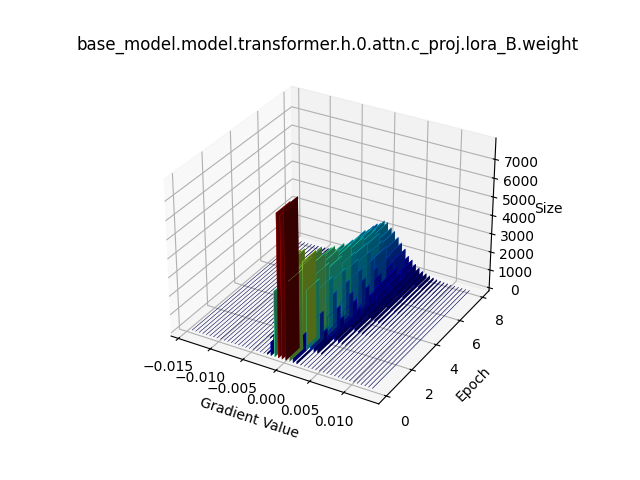}
     \end{subfigure}
     \begin{subfigure}[b]{0.24\textwidth}
         \centering
         \includegraphics[width=\textwidth]{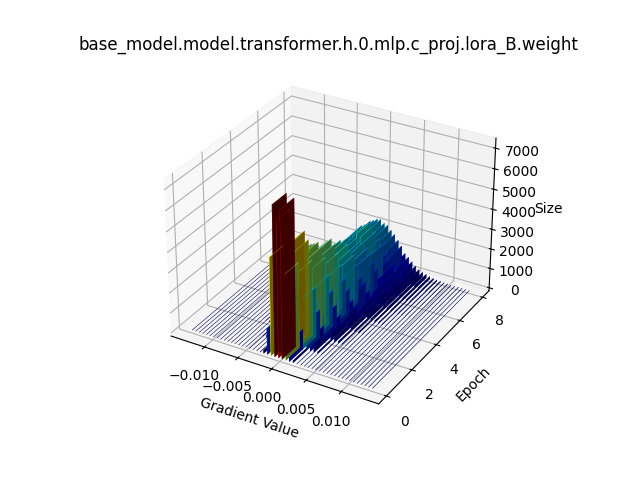}
     \end{subfigure}
     \begin{subfigure}[b]{0.24\textwidth}
         \centering
         \includegraphics[width=\textwidth]{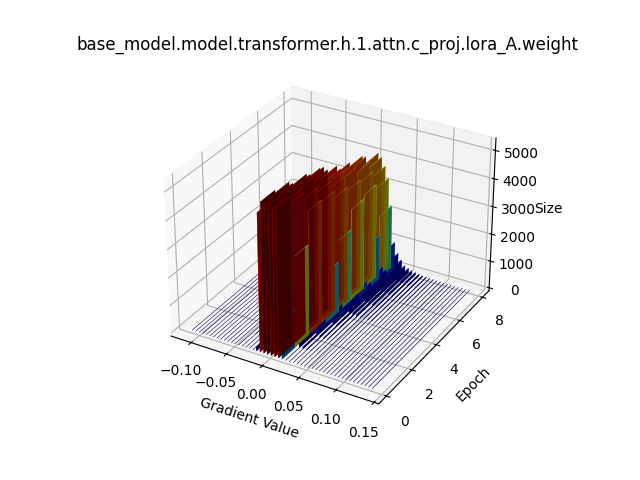}
     \end{subfigure}
     \begin{subfigure}[b]{0.24\textwidth}
         \centering
         \includegraphics[width=\textwidth]{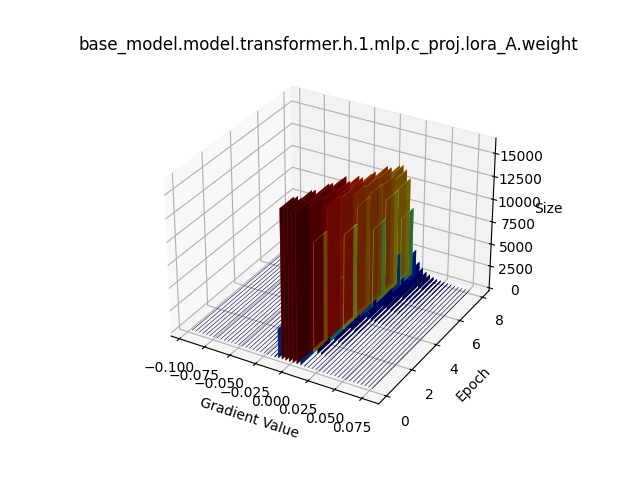}
     \end{subfigure}
     \begin{subfigure}[b]{0.24\textwidth}
         \centering
         \includegraphics[width=\textwidth]{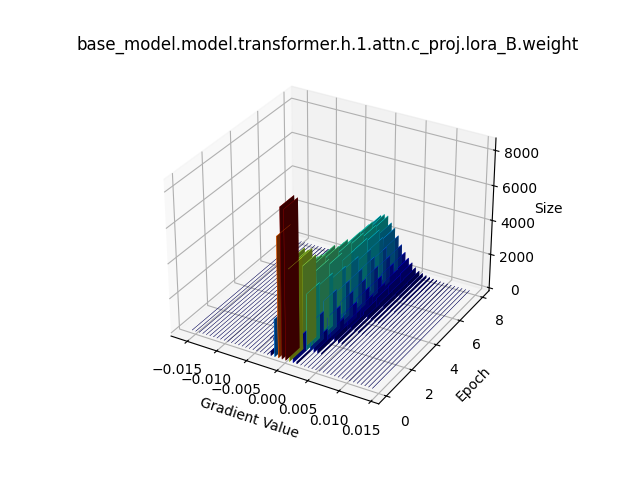}
     \end{subfigure}
     \begin{subfigure}[b]{0.24\textwidth}
         \centering
         \includegraphics[width=\textwidth]{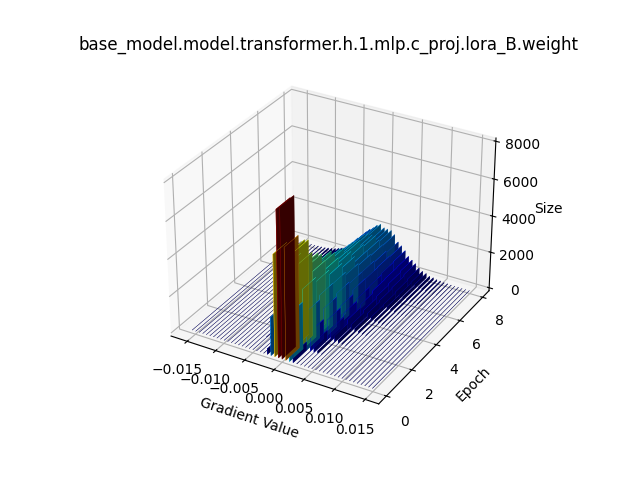}
     \end{subfigure}
     \begin{subfigure}[b]{0.24\textwidth}
         \centering
         \includegraphics[width=\textwidth]{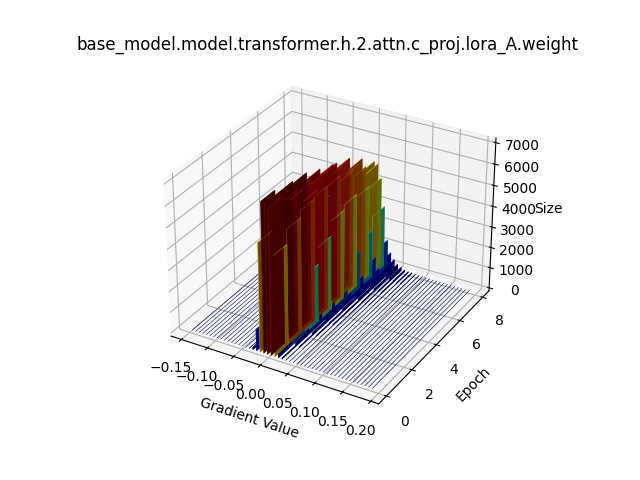}
     \end{subfigure}
     \begin{subfigure}[b]{0.24\textwidth}
         \centering
         \includegraphics[width=\textwidth]{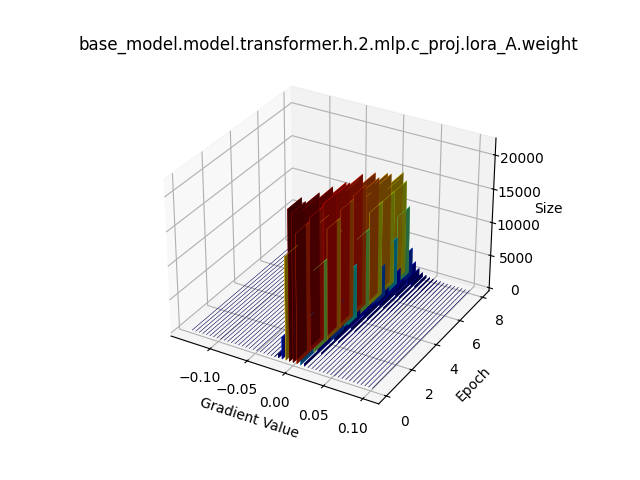}
     \end{subfigure}
     \begin{subfigure}[b]{0.24\textwidth}
         \centering
         \includegraphics[width=\textwidth]{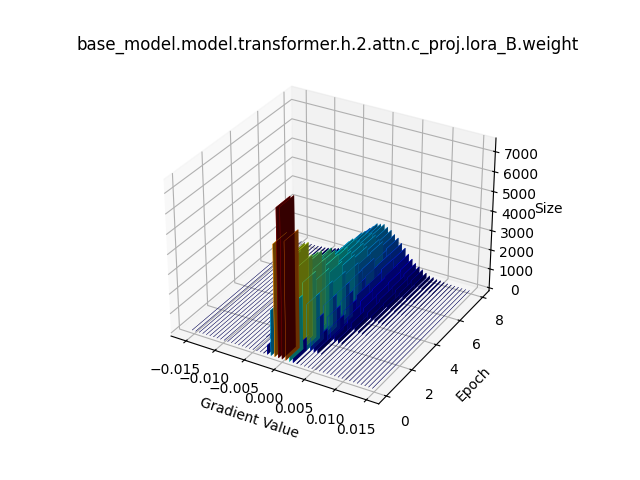}
     \end{subfigure}
     \begin{subfigure}[b]{0.24\textwidth}
         \centering
         \includegraphics[width=\textwidth]{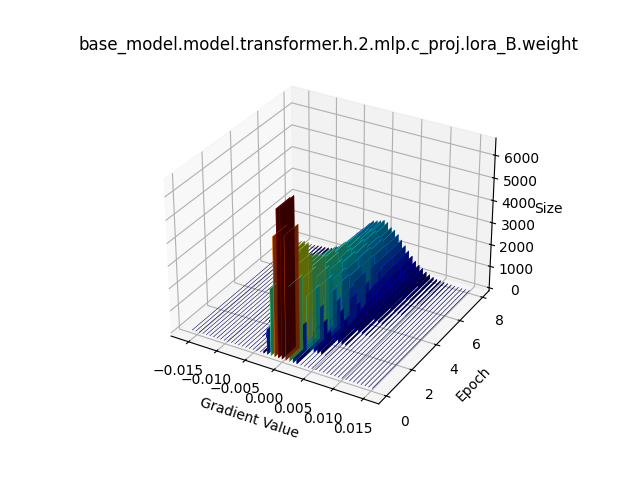}
     \end{subfigure}
     \begin{subfigure}[b]{0.24\textwidth}
         \centering
         \includegraphics[width=\textwidth]{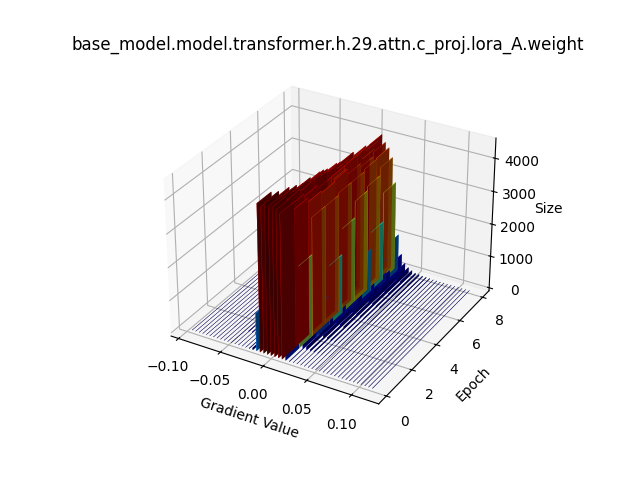}
     \end{subfigure}
     \begin{subfigure}[b]{0.24\textwidth}
         \centering
         \includegraphics[width=\textwidth]{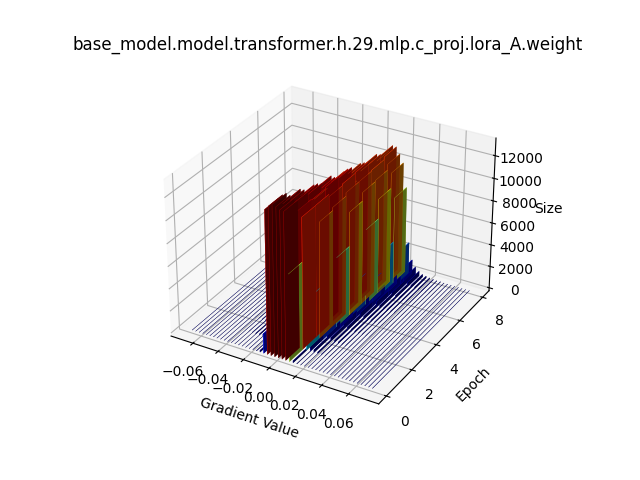}
     \end{subfigure}
     \begin{subfigure}[b]{0.24\textwidth}
         \centering
         \includegraphics[width=\textwidth]{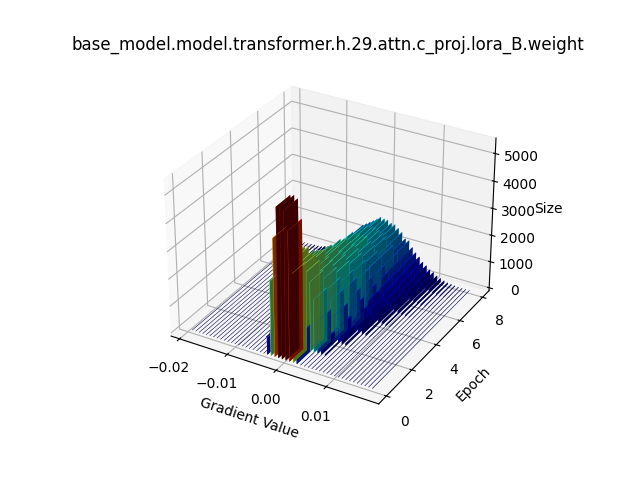}
     \end{subfigure}
     \begin{subfigure}[b]{0.24\textwidth}
         \centering
         \includegraphics[width=\textwidth]{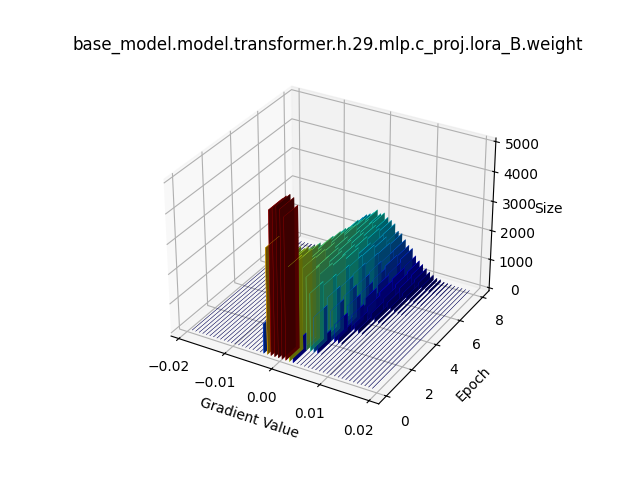}
     \end{subfigure}
     \begin{subfigure}[b]{0.24\textwidth}
         \centering
         \includegraphics[width=\textwidth]{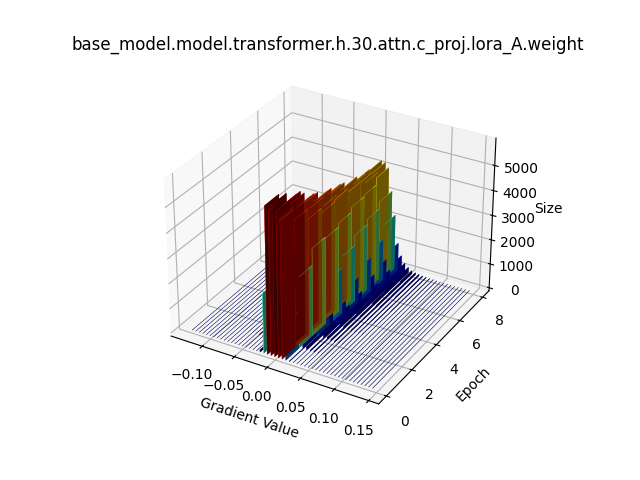}
     \end{subfigure}
     \begin{subfigure}[b]{0.24\textwidth}
         \centering
         \includegraphics[width=\textwidth]{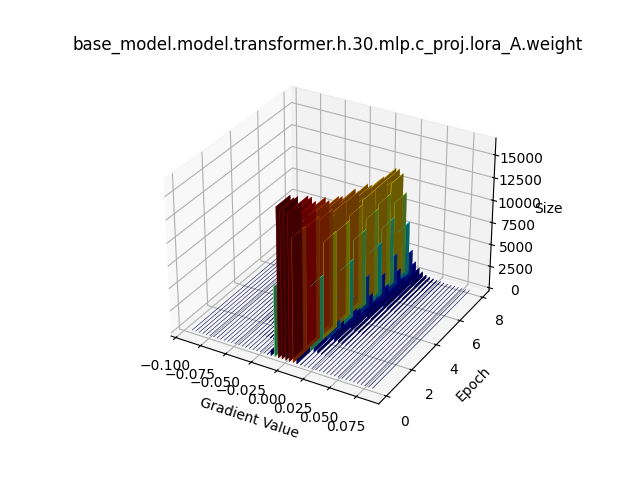}
     \end{subfigure}
     \begin{subfigure}[b]{0.24\textwidth}
         \centering
         \includegraphics[width=\textwidth]{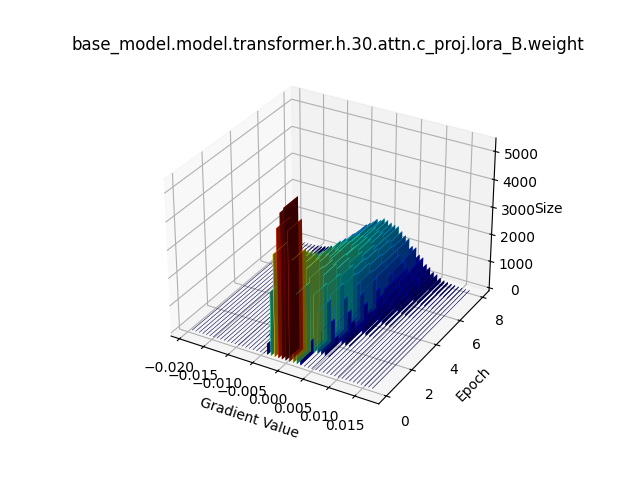}
     \end{subfigure}
     \begin{subfigure}[b]{0.24\textwidth}
         \centering
         \includegraphics[width=\textwidth]{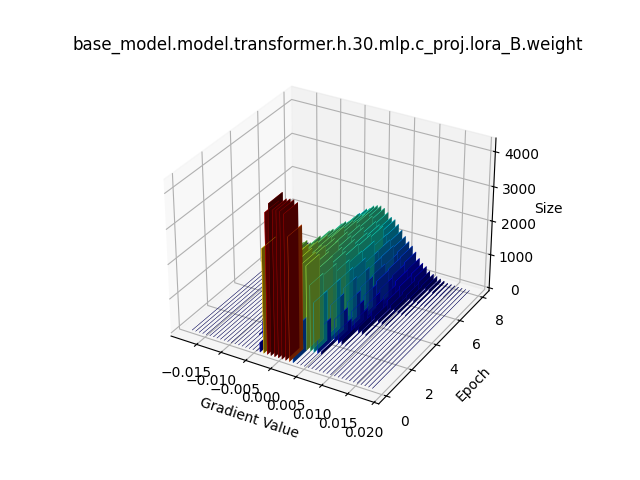}
     \end{subfigure}
     \begin{subfigure}[b]{0.24\textwidth}
         \centering
         \includegraphics[width=\textwidth]{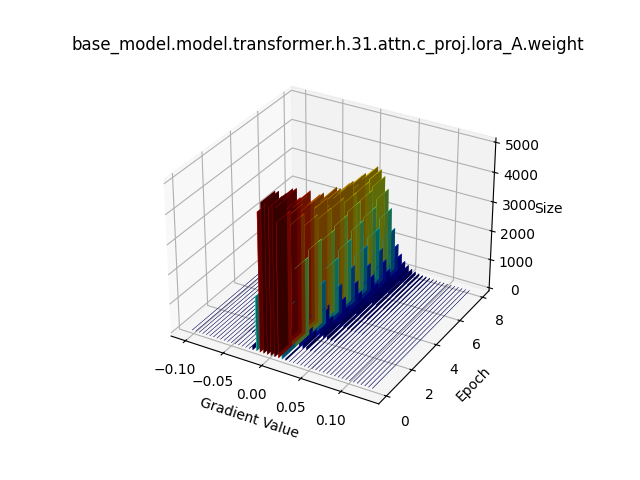}
     \end{subfigure}
     \begin{subfigure}[b]{0.24\textwidth}
         \centering
         \includegraphics[width=\textwidth]{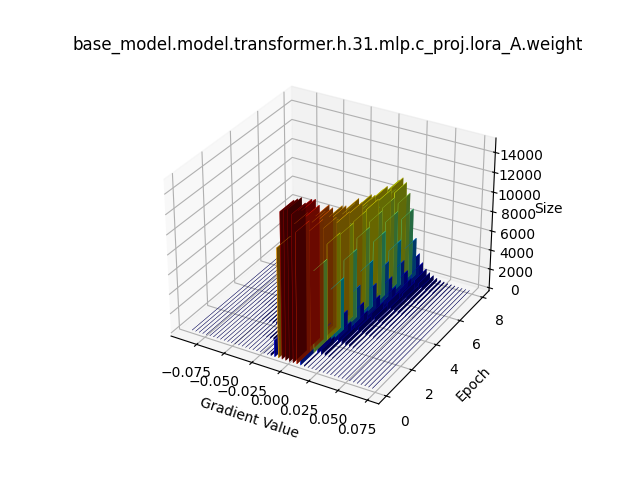}
     \end{subfigure}
     \begin{subfigure}[b]{0.24\textwidth}
         \centering
         \includegraphics[width=\textwidth]{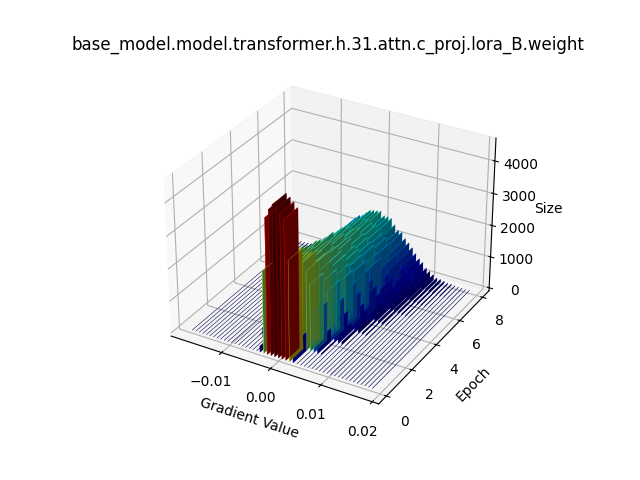}
     \end{subfigure}
     \begin{subfigure}[b]{0.24\textwidth}
         \centering
         \includegraphics[width=\textwidth]{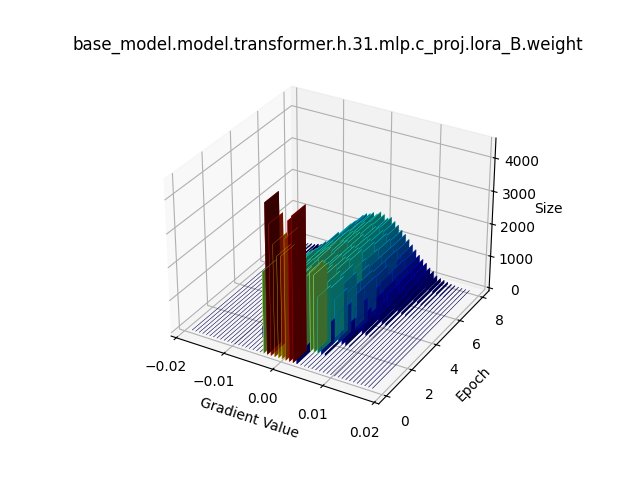}
     \end{subfigure}
\caption{Gradient Dynamical Visualization of Layer $\{0,1,2,29,30,31\}$.}
\label{fig:A_dist}
\end{figure*}

\begin{figure*}[htbp]
     \begin{subfigure}[b]{0.24\textwidth}
         \centering
         \includegraphics[width=\textwidth]{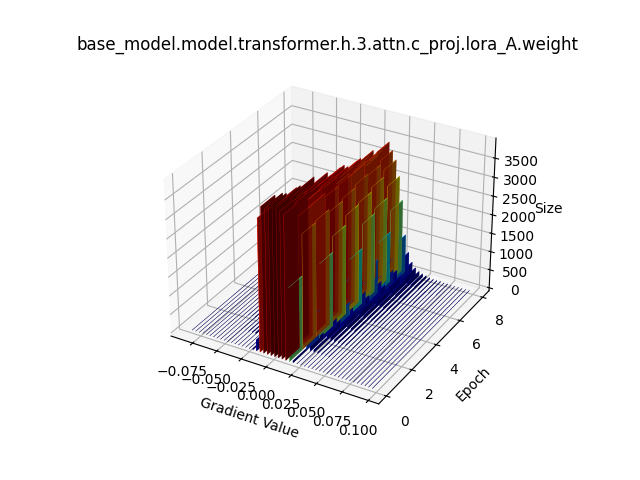}
     \end{subfigure}
     \begin{subfigure}[b]{0.24\textwidth}
         \centering
         \includegraphics[width=\textwidth]{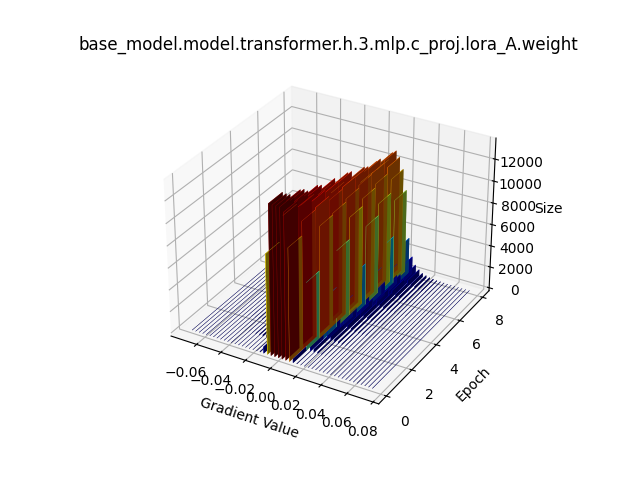}
     \end{subfigure}
     \begin{subfigure}[b]{0.24\textwidth}
         \centering
         \includegraphics[width=\textwidth]{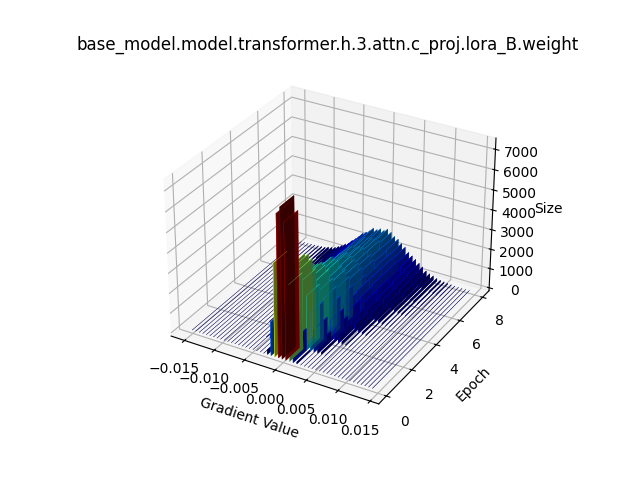}
     \end{subfigure}
     \begin{subfigure}[b]{0.24\textwidth}
         \centering
         \includegraphics[width=\textwidth]{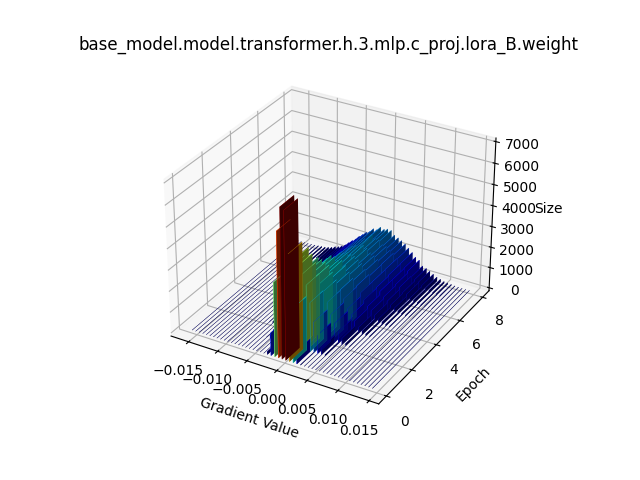}
     \end{subfigure}
     \begin{subfigure}[b]{0.24\textwidth}
         \centering
         \includegraphics[width=\textwidth]{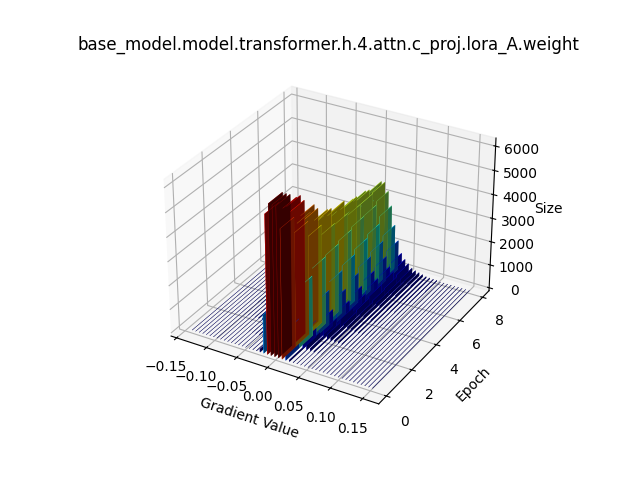}
     \end{subfigure}
     \begin{subfigure}[b]{0.24\textwidth}
         \centering
         \includegraphics[width=\textwidth]{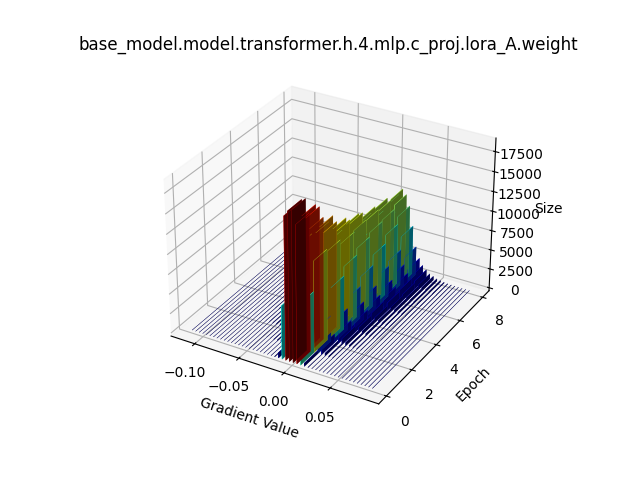}
     \end{subfigure}
     \begin{subfigure}[b]{0.24\textwidth}
         \centering
         \includegraphics[width=\textwidth]{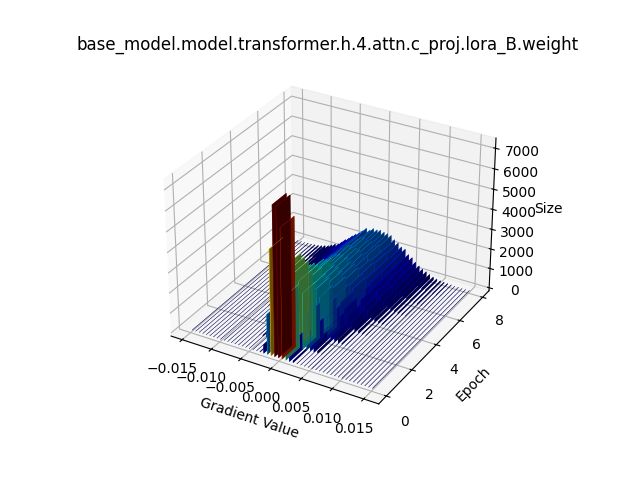}
     \end{subfigure}
     \begin{subfigure}[b]{0.24\textwidth}
         \centering
         \includegraphics[width=\textwidth]{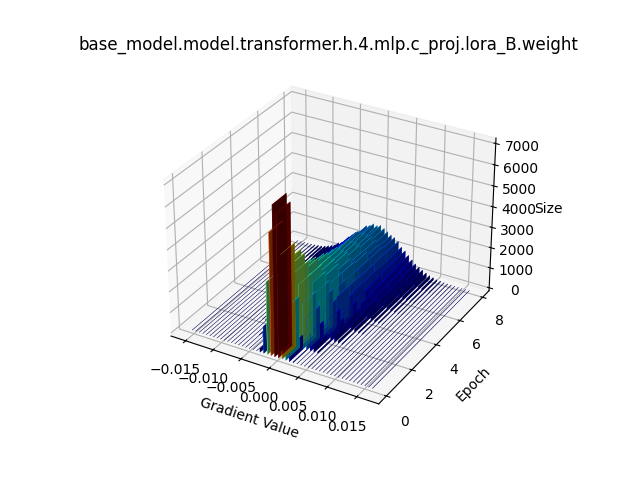}
     \end{subfigure}
     \begin{subfigure}[b]{0.24\textwidth}
         \centering
         \includegraphics[width=\textwidth]{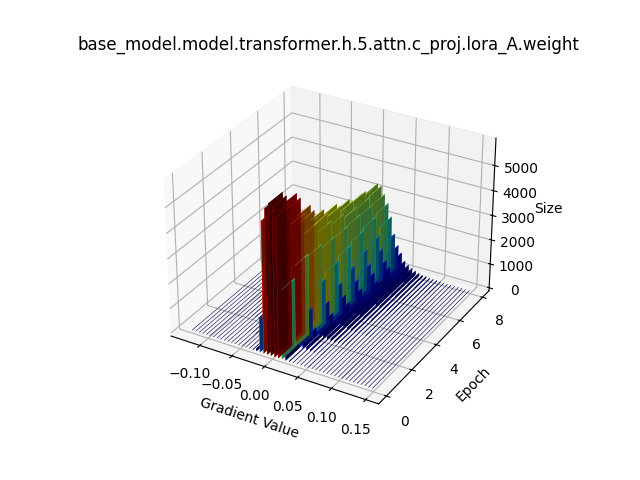}
     \end{subfigure}
     \begin{subfigure}[b]{0.24\textwidth}
         \centering
         \includegraphics[width=\textwidth]{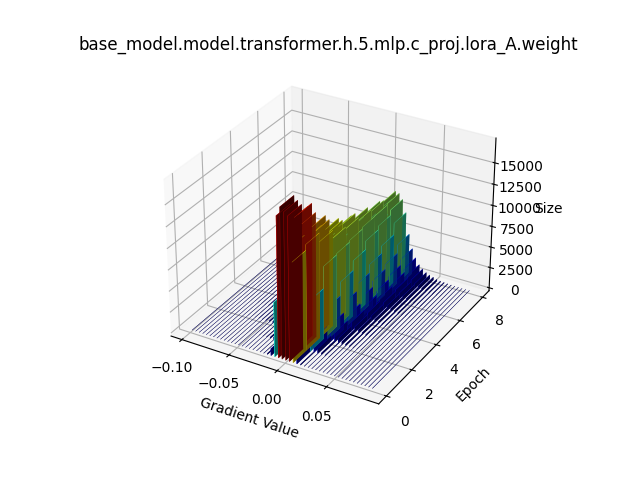}
     \end{subfigure}
     \begin{subfigure}[b]{0.24\textwidth}
         \centering
         \includegraphics[width=\textwidth]{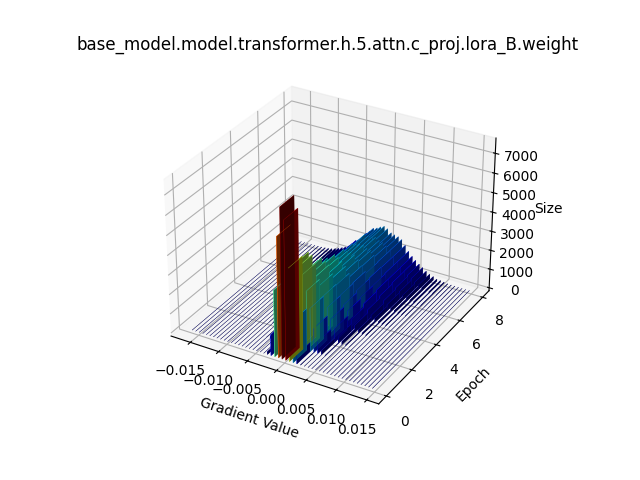}
     \end{subfigure}
     \begin{subfigure}[b]{0.24\textwidth}
         \centering
         \includegraphics[width=\textwidth]{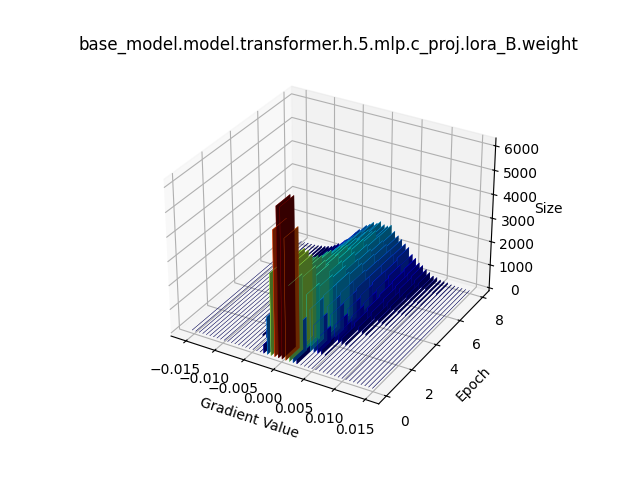}
     \end{subfigure}
     \begin{subfigure}[b]{0.24\textwidth}
         \centering
         \includegraphics[width=\textwidth]{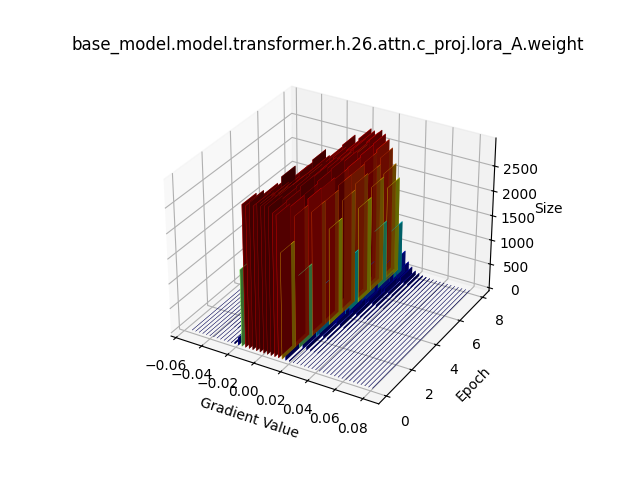}
     \end{subfigure}
     \begin{subfigure}[b]{0.24\textwidth}
         \centering
         \includegraphics[width=\textwidth]{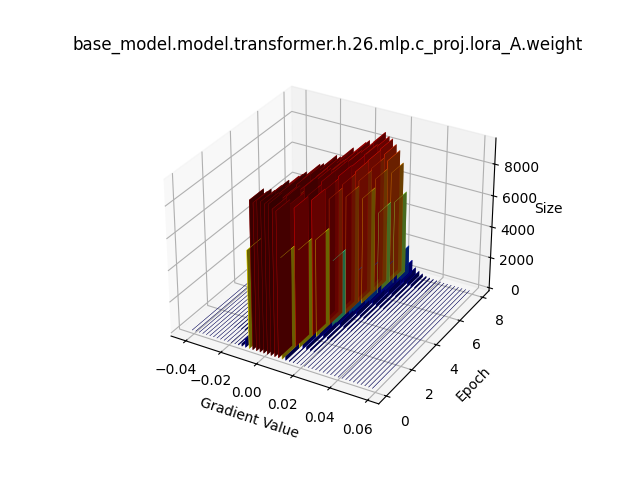}
     \end{subfigure}
     \begin{subfigure}[b]{0.24\textwidth}
         \centering
         \includegraphics[width=\textwidth]{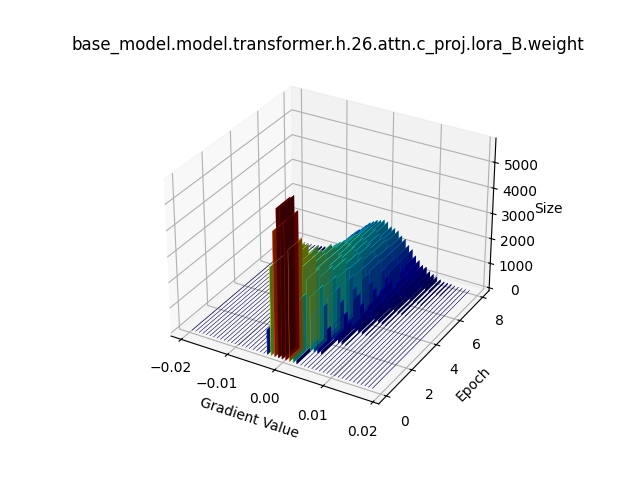}
     \end{subfigure}
     \begin{subfigure}[b]{0.24\textwidth}
         \centering
         \includegraphics[width=\textwidth]{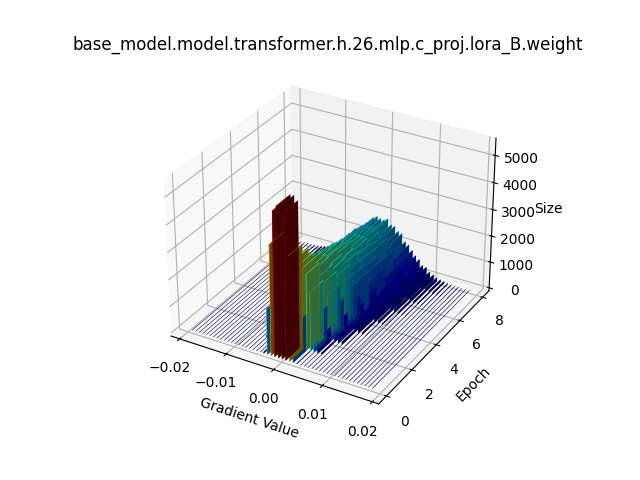}
     \end{subfigure}
     \begin{subfigure}[b]{0.24\textwidth}
         \centering
         \includegraphics[width=\textwidth]{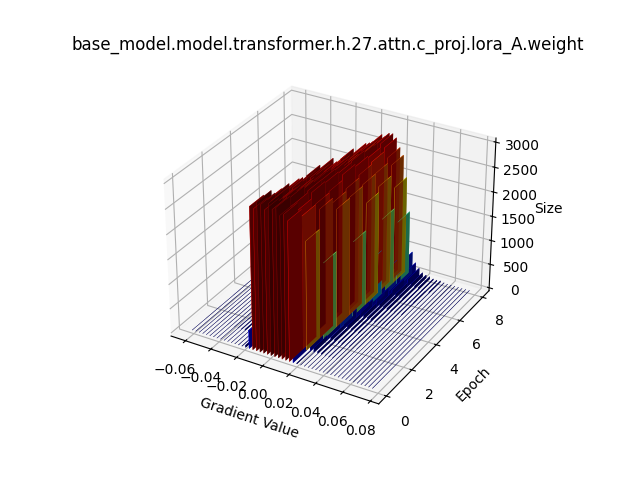}
     \end{subfigure}
     \begin{subfigure}[b]{0.24\textwidth}
         \centering
         \includegraphics[width=\textwidth]{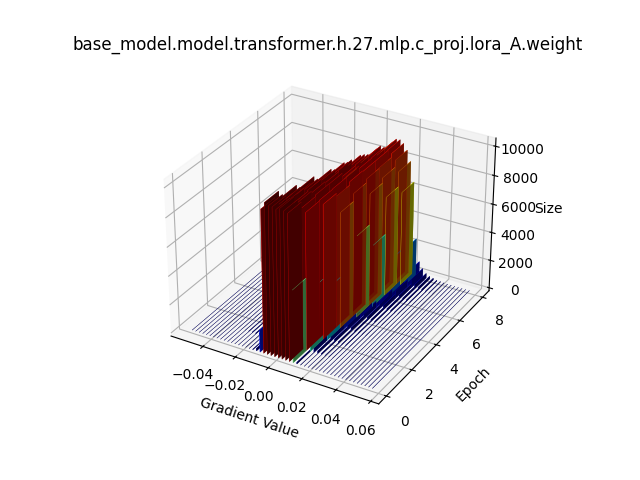}
     \end{subfigure}
     \begin{subfigure}[b]{0.24\textwidth}
         \centering
         \includegraphics[width=\textwidth]{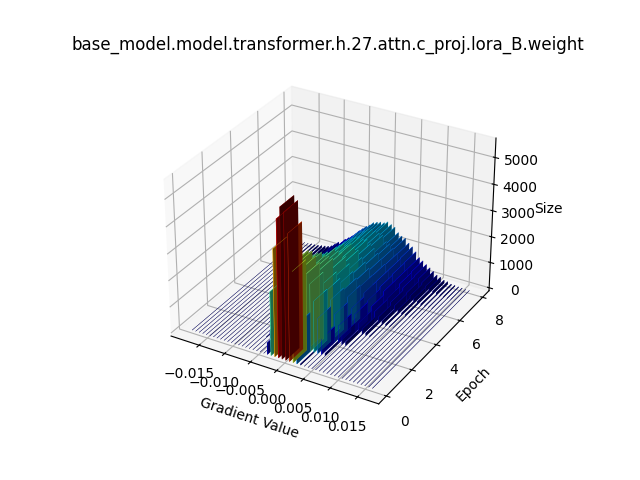}
     \end{subfigure}
     \begin{subfigure}[b]{0.24\textwidth}
         \centering
         \includegraphics[width=\textwidth]{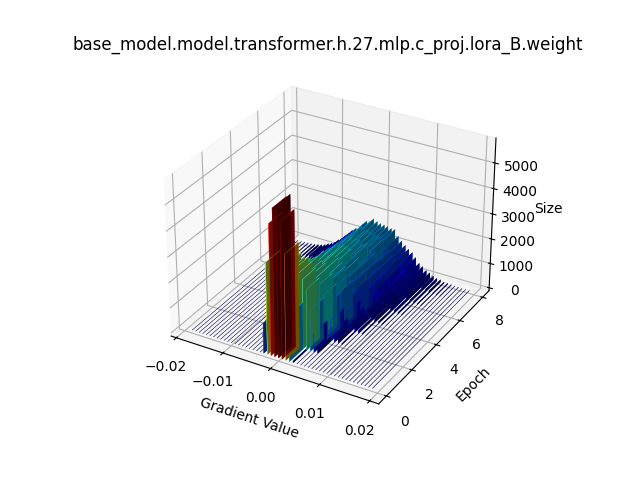}
     \end{subfigure}
     \begin{subfigure}[b]{0.24\textwidth}
         \centering
         \includegraphics[width=\textwidth]{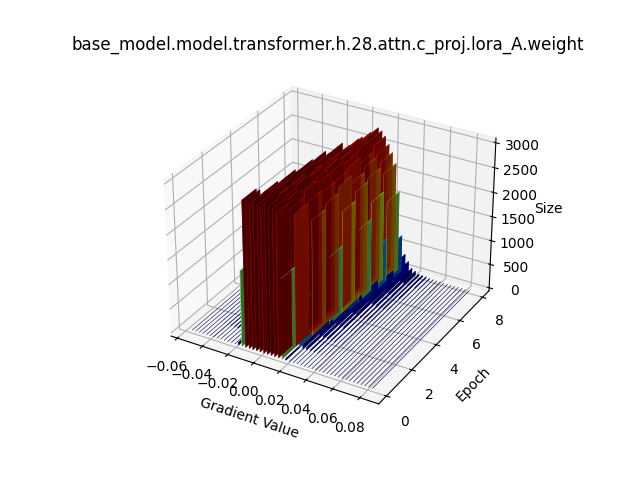}
     \end{subfigure}
     \begin{subfigure}[b]{0.24\textwidth}
         \centering
         \includegraphics[width=\textwidth]{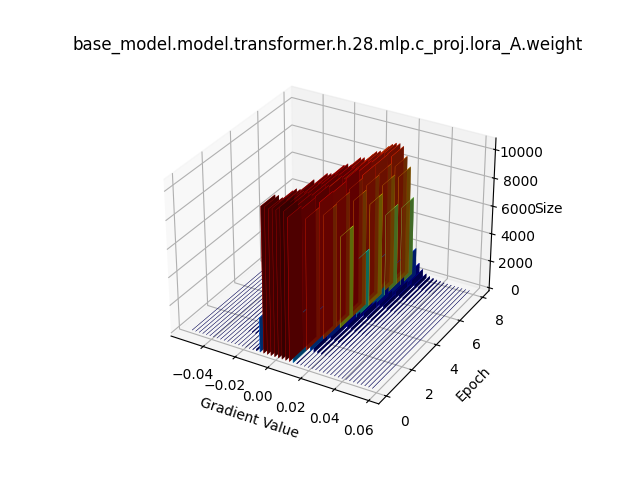}
     \end{subfigure}
     \begin{subfigure}[b]{0.24\textwidth}
         \centering
         \includegraphics[width=\textwidth]{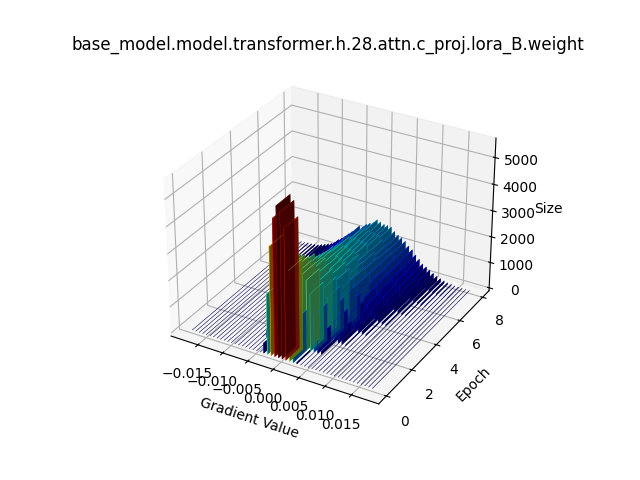}
     \end{subfigure}
     \begin{subfigure}[b]{0.24\textwidth}
         \centering
         \includegraphics[width=\textwidth]{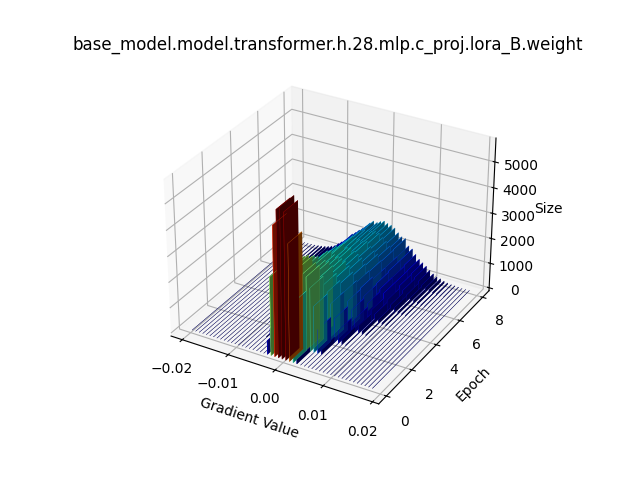}
     \end{subfigure}
\caption{Gradient Dynamic Visualization for Layer $\{4,5,6,26,27,28\}$.}
\label{fig:A_dist_2}
\end{figure*}

In this section, we present the distribution of the gradient throughout the learning epochs. Figure~\ref{fig:A_dist} illustrates the behavior of the low-rank decomposition matrices $\mathbf{A}$ and $\mathbf{B}$ across various layers during the training iterations. The x-axis shows the ranges of values of the matrices, the y-axis indicates the iteration steps, and the z-axis depicts the frequency of these values. Vertically, from top to bottom, the distributions of $\mathbf{A}$ and $\mathbf{B}$ are displayed in the layers $\{0,1,2,29,30,31\}$. Horizontally, from left to right, the distributions of $\mathbf{A}$ in the attention layer and the Multi-Layer Perceptron (MLP) and $\mathbf{B}$ in the attention layer and the MLP layer are shown. In Figure~\ref{fig:A_dist_2}, we visualize the gradients for layers $\{4,5,6,26,27,28\}$. The visualization reveals that the gradient flattens as the learning epoch progresses. A significant difference is observed between the gradient distributions of the $\mathbf{A}$ and $\mathbf{B}$ parts.

\end{document}